\PassOptionsToPackage{english}{babel}
\documentclass{article}

\usepackage{microtype}
\usepackage{graphicx}
\usepackage{subcaption}
\usepackage{booktabs} 
\usepackage{pifont}    
\usepackage{xcolor}    
\usepackage{multirow}  

\usepackage{hyperref}

\usepackage[preprint]{icml2026}

\usepackage{amsmath}
\usepackage{amssymb}
\usepackage{mathtools}
\usepackage{amsthm}
\usepackage{xurl}

\usepackage[capitalize,noabbrev]{cleveref}

\theoremstyle{plain}

\theoremstyle{definition}

\theoremstyle{remark}

\usepackage[textsize=tiny]{todonotes}

\icmltitlerunning{Beyond Via: Analysis and Estimation of the Impact of Large Language Models in Academic Papers}

\begin{document}

\twocolumn[
  \icmltitle{Beyond Via: Analysis and Estimation of the Impact of Large Language Models in Academic Papers}



  \icmlsetsymbol{equal}{*}

    \begin{icmlauthorlist}
    \icmlauthor{Mingmeng Geng}{equal,ens,lattice}
    \icmlauthor{Yuhang Dong}{equal,fau}
    \icmlauthor{Thierry Poibeau}{ens,lattice}
    \end{icmlauthorlist}
    
    \icmlaffiliation{ens}{École Normale Supérieure (ENS) – Université Paris Sciences et Lettres (PSL)}
    \icmlaffiliation{lattice}{Laboratoire Lattice}
    \icmlaffiliation{fau}{Friedrich-Alexander-Universität Erlangen-Nürnberg (FAU)}
    
    \icmlcorrespondingauthor{Mingmeng Geng}{mingmeng.geng@ens.psl.eu}

  \icmlkeywords{Machine Learning, ICML}

  \vskip 0.3in
]



\printAffiliationsAndNotice{}  

\begin{abstract}

Through an analysis of arXiv papers, we report several shifts in word usage that are likely driven by large language models (LLMs) but have not previously received sufficient attention, such as the increased frequency of ``\textbf{beyond}'' and ``\textbf{via}'' in titles and the decreased frequency of ``\textbf{the}'' and ``\textbf{of}'' in abstracts. Due to the similarities among different LLMs, experiments show that current classifiers struggle to accurately determine which specific model generated a given text in multi-class classification tasks. Meanwhile, variations across LLMs also result in evolving patterns of word usage in academic papers. By adopting a direct and highly interpretable linear approach and accounting for differences between models and prompts, we quantitatively assess these effects and show that real-world LLM usage is heterogeneous and dynamic.\footnote{Visualization of word usage patterns in arXiv abstracts: \\ \url{https://llm-impact.github.io/}}
\end{abstract} 

\section{Introduction}

The increasing impact of large language models (LLMs) in academic publications has been observed~\cite{liang2024monitoring,geng2024chatgpt,kobak2024delving}. As LLMs continue to develop, has their impact evolved more recently?

\begin{figure}[!t]
    \begin{subfigure}{\columnwidth}
        \centering
        \includegraphics[width=0.48\columnwidth]{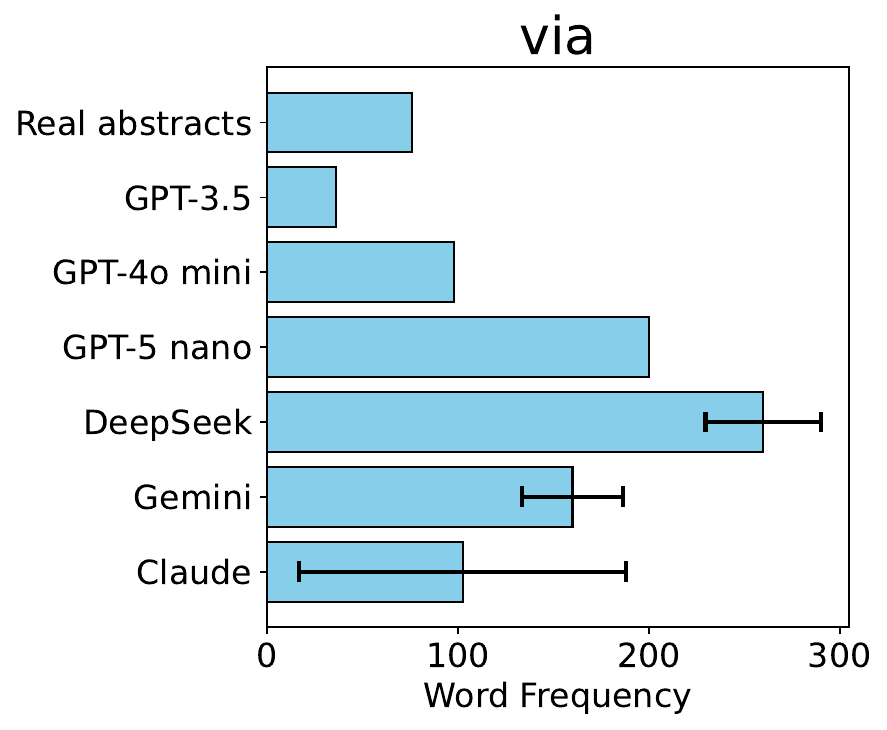}
        \includegraphics[width=0.48\columnwidth]{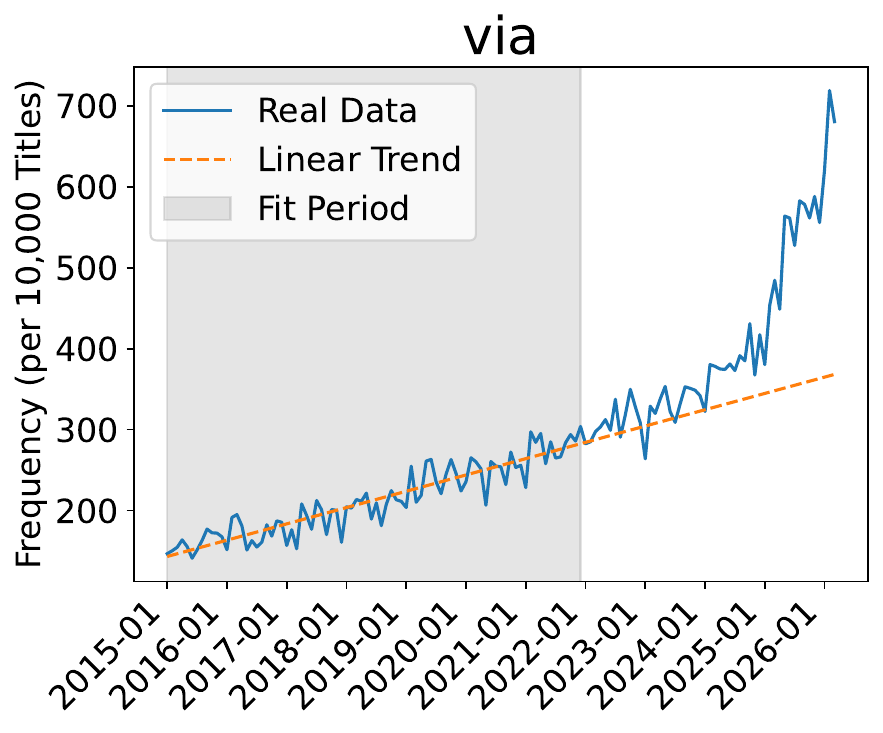}
        \caption{In arXiv paper titles.}
        \label{via}
    \end{subfigure}
    \begin{subfigure}{\columnwidth}
        \centering
        \includegraphics[width=0.48\columnwidth]{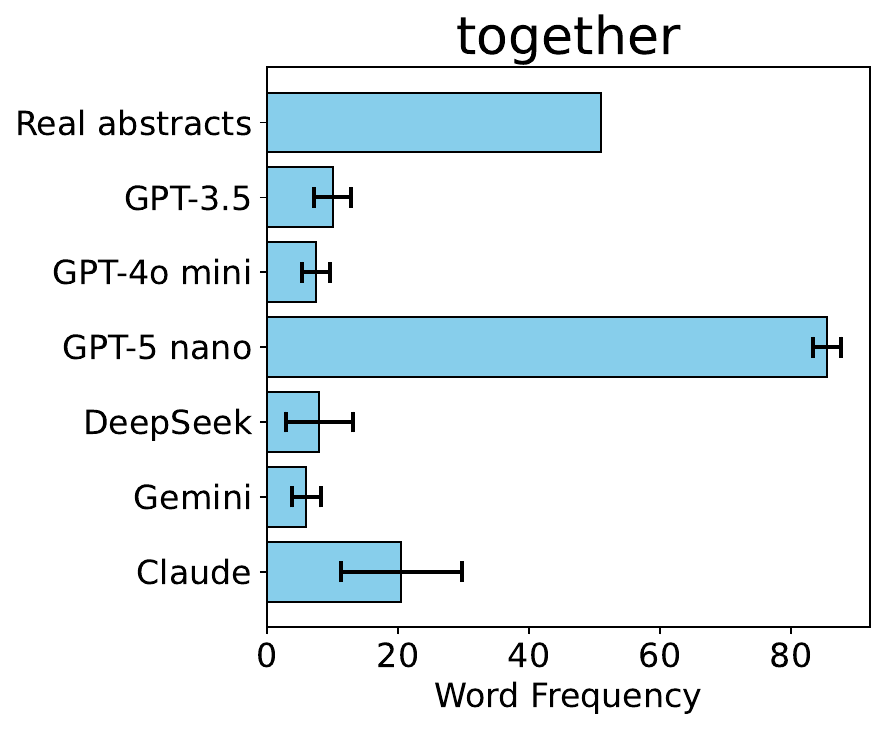}
        \includegraphics[width=0.48\columnwidth]{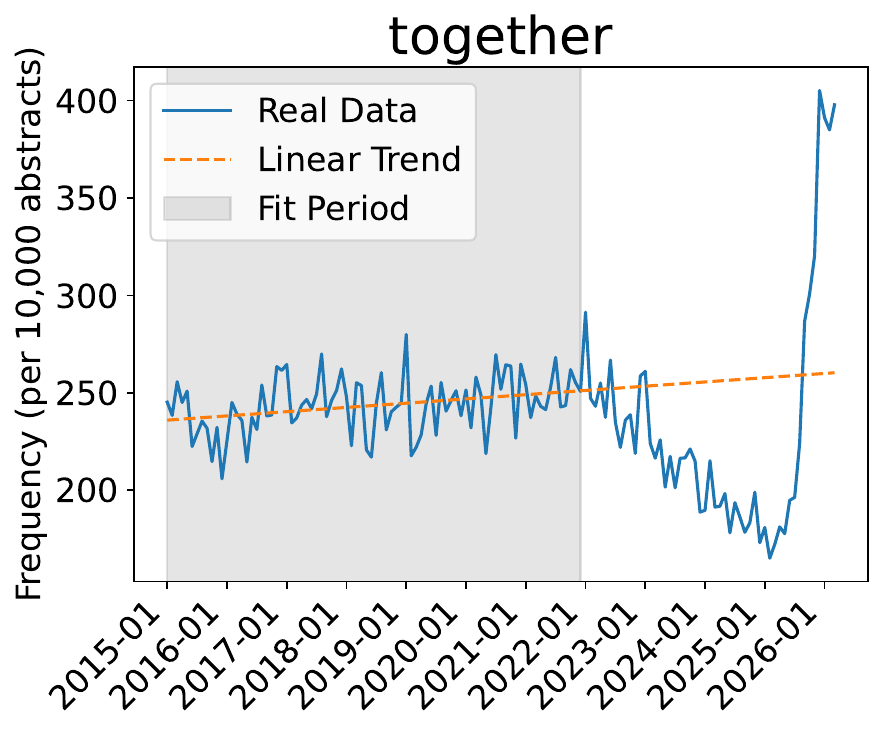}
        \includegraphics[width=0.95\columnwidth]{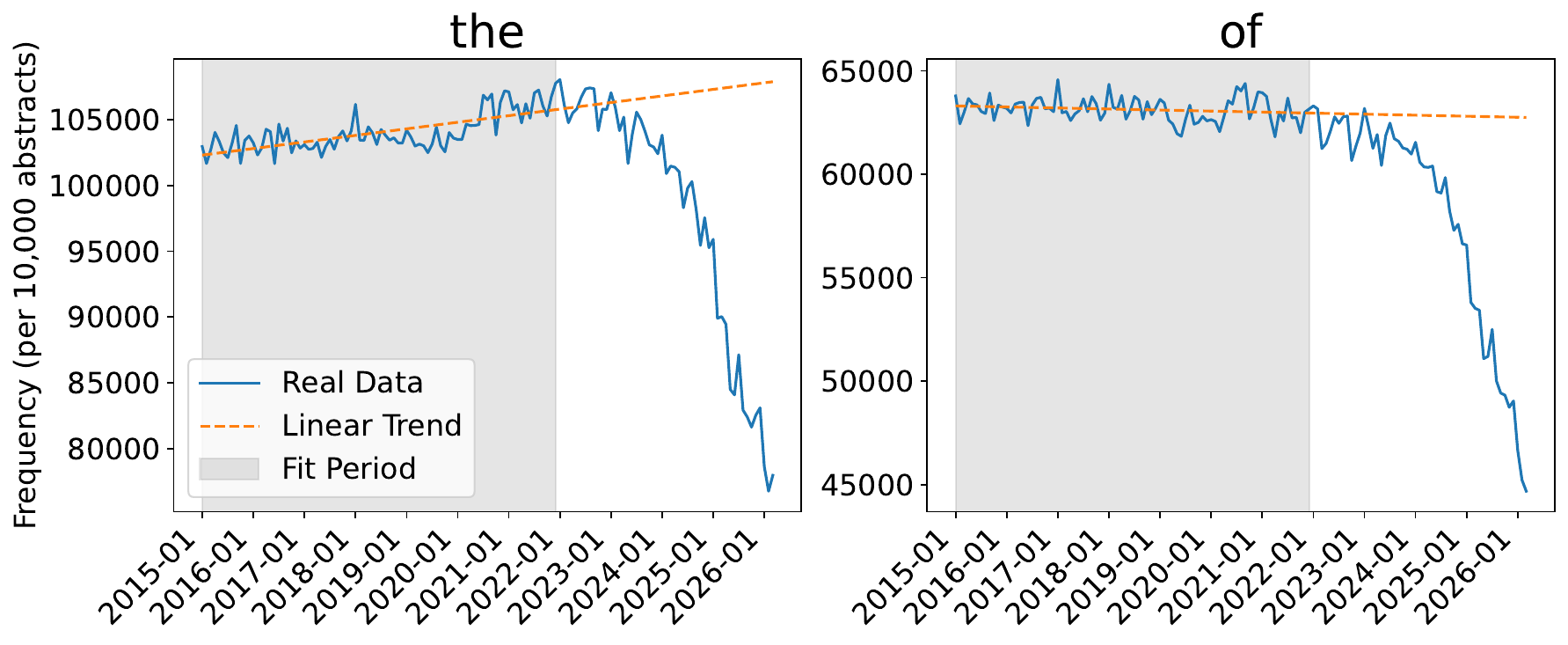}
        \caption{In arXiv paper abstracts.}
        \label{together_the_of}
    \end{subfigure}
    \caption{\textbf{Top-left \& Middle-left}: Word frequency comparison for generated titles or rewritten abstracts based on 2,000 real arXiv abstracts; error bars denote variance across models and prompts. \textbf{Remaining panels}: Temporal trends of word frequencies in real arXiv data, with the yellow dashed line fitted on data from 2015 to 2021 and extended to early 2026.}
    \label{examples}
    \vspace{-1em}
\end{figure}

For example, some researchers might have begun to reduce their use of certain LLM-style (more precisely, ChatGPT-style) expressions, such as ``\textbf{delve}'', around late 2023 and early 2024~\cite{geng2025human}. Different LLMs possess their own idiosyncrasies, which can be used to classify the text they generate~\cite{wu2023llmdet,antoun2024text,sun2025idiosyncrasies}. Considering the multiple updates to ChatGPT and the emergence of other models, this paper aims to analyze and estimate the impact of LLMs on academic publications in relation to these developments.

As depicted in Figure~\ref{via}, certain LLMs have a preference for the word ``\textbf{via}'' when generating titles from real arXiv abstracts, whose frequency is also becoming more common in real arXiv paper titles from 2025. Meanwhile, Figure~\ref{together_the_of} shows that the frequency of ``\textbf{together}'' first declines markedly and is then followed by a rapid increase, likely reflecting the contrasting preferences between newer and older LLMs regarding this term. Moreover, ``\textbf{the}'' and ``\textbf{of}'', two of the most commonly used words in many English corpora, have experienced a clear decline in frequency within arXiv abstracts.

The use of words and expressions in academic writing is constantly evolving~\cite{bizzoni2020linguistic,wang2023linguistic,cheng2024anthroscore}. Researchers are also continuously monitoring the use of specific indicative words to estimate the usage of LLMs in academic writing~\cite{gray2025estimating,he2025academic}. Comparisons between texts generated by LLMs and those written by humans can now be carried out using different techniques~\cite{wu2025survey}. 

In this work, we first collect arXiv papers based on a publicly available dataset and use different LLMs to simulate parts of the abstracts submitted before the emergence of ChatGPT, either by rewriting the abstracts or generating titles. We then compare the similarities and differences between the texts produced by LLMs and those written by humans from different perspectives, as well as the variations between texts generated by different LLMs. Although models have been advancing, their outputs still differ in some way from human-written texts. Finally, we choose to analyze and estimate the impact of LLMs in academic publications through word usage, since different LLMs favor different words and word frequency is a relatively interpretable measure. 

Our findings suggest that the influence of LLMs on academic publications is growing, and the variety and the development of LLMs has led to further variations. The tools and metrics for analyzing and monitoring their impact need to evolve as well. Although more sophisticated methods may provide numerically more accurate estimates, direct approaches can also offer new insights into the understanding and visualizing of the societal impact of LLMs. 

\section{Data}
\label{sec:data}

\subsection{arXiv Data}

This paper analyzes abstracts of arXiv papers, based on a dataset that is updated weekly on Kaggle~\cite{arxiv_org_submitters_2024}\footnote{\url{https://www.kaggle.com/datasets/Cornell-University/arxiv}}. These data include more than 2.9 million arXiv papers across different fields. Considering the quantity and rapid update rate of arXiv papers, they may well be the best avenue for timely monitoring of LLM influence on academic writing.

As our study requires analyzing how effects vary over time, we need to incorporate timestamps into the analysis. In this dataset, two such timestamps are available: the paper ID \texttt{id} (indicating the initial submission) and the last update time \texttt{update\_date} (indicating the most recent revision).

\subsection{Data Preparation}

Dataset versions from version 1 (released in August 2020) to version 277 (released in March 2026) were collected, with only newly added papers retained in each version. This means that the time implied by the arXiv ID is essentially the submission time, and even with revisions, the delay will not exceed a week. In this way, we can reduce the temporal delay introduced by paper updates.

We categorize papers based on arXiv categories, considering only the first category for papers with multiple categories. We also apply \texttt{re.findall(r'\textbackslash b\textbackslash w+\textbackslash b', x.lower())} in Python to strip punctuation and convert all text to lowercase.

Based on their update dates in version 265, we randomly selected 2,000 papers from January to October 2022 to simulate using LLMs, i.e., prior to the release of ChatGPT.

\subsection{Other Data}

Only frequent words are considered to estimate the impact of LLMs. In addition to word frequencies in the arXiv dataset, we also used data from the Google Books Ngram dataset to establish a threshold for selecting common words. We are also interested in how stopwords are used in LLMs, with reference to the list in Natural Language Toolkit (NLTK)~\cite{bird2006nltk}. 

\section{Methods}

\subsection{Trend of Word Frequency Over Time}
We denote the observed frequency of word $w$ in the dataset at time $t$ by $f^{\mathrm{d}}_w(t)$, which is defined by the following equation:
\begin{equation}
    f^{\mathrm{d}}_w(t) = f_w(t) + \varepsilon_w(t) \label{f_d}
\end{equation}
where $f_w(t)$ refers to the pattern of how word frequency changes over time, while $\varepsilon_w(t)$ represents random variations. 

Even without considering randomness, the frequency of words can be influenced by multiple factors, such as changes in research topics or writing conventions. In the following, we assume that the baseline word frequency changes linearly over time and then model the effect of LLM usage as a factor that modifies this established linear trend, as shown in the formula below:
\begin{equation}
    f_w(t) = a_w + b_w t + c_w(t)
\end{equation}
where $a_w$ and $b_w$ are word-specific coefficients, while $c_w(t)$ represents the effect introduced by LLMs.

Based on observational data from the pre-LLM period $t \in \mathcal{T}_1$, we estimate the parameters $a_w$ and $b_w$ and $\hat{a}_w$ and $\hat{b}_w$. These estimates are then used to predict the word frequency $f^{\mathrm{pred}}_w(t)$ over the later time interval $t \in \mathcal{T}_2$ as follow:
\begin{equation}
    f^{\mathrm{pred}}_w(t) = \hat{a}_w + \hat{b}_w t \, . \label{f_p}
\end{equation}

As shown in Figures~\ref{examples}, the linear regression results based on data from 2021 and earlier can also reasonably predict the word frequencies for the first ten months of 2022, i.e., before the emergence of ChatGPT. Therefore, we can use the linear regression described above to predict how word frequencies would change over time as if LLMs did not exist.

\subsection{Impact Estimation}

For different models $m$, they have varying preferences for words, which also depend on the prompt. Let $f^0_w$ represent the word frequency in human-written text. For model $m$ under prompt $p$, the word frequency is $f^{m,p}_w$, and its mean across prompts is denoted by $f^m_w$.

For the observed values in Equation~\ref{f_d} and the predicted values in Equation~\ref{f_p}, we can easily obtain the following ratio:
\begin{equation}
r_w(t) = \frac{f^{\mathrm{d}}_w(t)}{f^{\mathrm{pred}}_w(t)} \, .
\end{equation}

If we further assume that $\eta_0(t)$ represents the proportion in human-written text at time $t$, and $\eta_{m,p}(t)$ represents the proportion of text generated by model $m$ and prompt $p$. Then, under the assumption that the observed text is a mixture of these components, we can obtain the following approximate relationship:
\begin{equation}
r_w(t) \approx \eta_0(t) + \sum_{m \in \mathcal{M}, p \in \mathcal{P}} \eta_{m,p}(t) \frac{f_w^{m}}{f_w^{0}},
\end{equation}
where $\mathcal{M}$ represents the set of all LLMs under consideration, and $\mathcal{P}$ represents the set of all prompts being considered.

If we denote all the unknowns related to $\eta$ as the vector $\boldsymbol{\eta}(t)$, and define $\mathcal{M^*}$ and $\mathcal{P^*}$ to include both human and non-prompted cases to accommodate $\eta_0$. Then for the set of all considered words $\mathcal{W}$, $\boldsymbol{\eta}(t)$ can be determined using the following relation:
\begin{equation}
  \begin{aligned}
\min_{\boldsymbol{\eta}(t)} \quad &
\sum_{w \in \mathcal{W}}
\left(r_w(t)-\sum_{m \in \mathcal{M^*}, p \in \mathcal{P^*}} \eta_{m,p}(t)
\frac{f_w^{m,p}}{f_w^{0}}
\right)^2 \\[6pt]
\text{s.t.} \quad &
\eta_0(t) + \sum_{m \in \mathcal{M^*}, p \in \mathcal{P}} \eta_{m,p}(t) = 1, \\[4pt]
&  \eta_{m,p}(t) \in [\ell_{m,p}, u_{m,p}], \quad \forall m \in \mathcal{M^*}, \, p \in \mathcal{P^*}. \label{eq_eta_estimation}
\end{aligned}  
\end{equation}

It is important to note that the value of $\eta_0$ may be very close to 1, for example, before the emergence of ChatGPT. At the same time, other values of $\eta$ may be close to 0. Therefore, although the loss function can be adjusted, it needs to be done with caution.

\subsection{Text Similarity Comparison and Distinction}
We explore two different tasks: generating the title and modifying the abstract. To examine the similarities and differences between human-authored texts and those generated by LLMs, as well as across outputs from different LLMs, we initially assess textual similarity using multiple metrics. We then attempt to train classifiers to categorize texts generated by different LLMs.

For example, we consider three specific cases of ROUGE~\cite{lin2004rouge} to compare the lexical similarity of texts: ROUGE-1 (word-level comparison), ROUGE-2 (bigram-level comparison), and ROUGE-L (comparison based on the longest common subsequence). We also use BERTScore~\cite{zhang2019bertscore}, based on a pre-trained semantic neural network, to assess their semantic similarity between two pieces of text. 

Referring to the design and parameters of a previous study~\cite{sun2025idiosyncrasies}, we also train classifiers to distinguish texts generated by different LLMs, based on three classical models: BERT~\cite{devlin2019bert}, GPT-2~\cite{radford2019language}, T5~\cite{raffel2020exploring}, and LLM2Vec~\cite{behnamghader2024llm2vec}. 

The selection of parameters, such as the learning rate is consistent with that of the previous paper~\cite{sun2025idiosyncrasies}. Although adjusting the model and its parameters might yield better classification results, the purpose of this section is to validate our dataset using the current best open-source classifier, so no modifications were made.

\subsection{Word Frequency Comparison}
Word frequency can also be analyzed from various perspectives. For example, we can define frequency change ratio as follows:
\begin{equation}
    r_{w,m} =\frac{f^m_w-f^0_w}{f^0_w} \, . \label{r_w_m}
\end{equation}
The variance of the frequency of a word $w$ in the same texts given a prompt $p$ across different models $m$ is expressed as:
\begin{equation}
    \mathrm{Var}_{w,p} = \frac{1}{|\mathcal{M^*}|-1} \sum_{m \in \mathcal{M^*}} (f^{m,p}_{w} - \bar{f}^p_m)^2
\end{equation}
where the average value $\bar{f}^p_m$ can be represented as
\begin{equation}
    \bar{f}^p_m = \frac{1}{|\mathcal{M^*}|} \sum_{m \in \mathcal{M^*}} f^{m,p}_{w} \, .
\end{equation}
Subsequently, the coefficient of variation is defined as
\begin{equation}
    \mathrm{CV}_{w,p} = \frac{\sqrt{\mathrm{Var}_{w,p}}}{\bar{f}^p_m} \, . \label{cv}
\end{equation}
Therefore, we can identify which words behave differently across different LLMs.

\section{Simulation}
\subsection{Models}

\begin{table}[t]
    \caption{Release timeline of the LLMs used in this paper. Dates indicate the initial public availability, verified by official announcements.}
    \label{tab:llm_release_dates}
    \centering
    \begin{small} 
    \begin{tabular}{llr}
        \toprule
        \textbf{Model Name} & \textbf{Release Date} \\
        \midrule
        GPT-3.5 \citep{openai_chatgpt}       & Dec 30, 2022 \\
        GPT-4o mini \citep{openai_gpt4o_mini}& Jul 18, 2024 \\
        GPT-5 nano \citep{openai_gpt5_nano}  & Aug 07, 2025 \\ 
        \midrule
        DeepSeek V3 \citep{deepseek_v3}      & Dec 26, 2024 \\
        DeepSeek R1 \citep{deepseek_r1}      & Jan 20, 2025 \\
        DeepSeek V3.2 \citep{deepseek_v3_2}  & Dec 01, 2025 \\ 
        \midrule
        Gemini 2.5 Flash \citep{google_gemini_2_5} & Jun 17, 2025 \\
        Gemini 2.5 Pro \citep{google_gemini_2_5}   & Jun 17, 2025 \\
        Gemini 3 Flash \citep{google_gemini_3}     & Dec 17, 2025 \\
        \midrule
        Claude 3 Haiku \citep{anthropic_claude_3_haiku}     & Mar 13, 2024 \\
        Claude Haiku 4.5 \citep{anthropic_claude_haiku_4_5} & Oct 15, 2025 \\
        \bottomrule
    \end{tabular}
    \end{small}
    \vspace{-1em}
\end{table}

Simulations were conducted using nine different models from the GPT, DeepSeek, Gemini, and Claude for simulation. Their release dates range from November 2022 to December 2025, as illustrated in Table~\ref{tab:llm_release_dates}. More details on the simulation using the API can be found in Appendix Section~\ref{api_usage}.

We are aware that some LLMs are not covered in this study, and some versions of GPT and Claude are not included in the experiments. But the models we choose are expected to yield fairly stable estimation results.

\subsection{Prompts}
\label{prompts}

Given the diversity of real-world application scenarios, we adopt two prompts of different lengths, one long and one short, for simulation.

\textbf{Short Prompt for Abstract Rewriting :} It is possible that authors restrict their use of LLM tools to light assistance, such as minor rewrites. Therefore, we consider the short prompt as follows.

\begin{quote}
    \small\ttfamily
    Revise the following sentences. Please only output the revised text in JSON format. Only one version. No explanations. Only plain text.\\
    EXAMPLE JSON OUTPUT: \{"text": "Your revised version of the provided sentences goes here."\}
\end{quote}

\textbf{Other Prompts:} We can also provide more details in a longer prompt to simulate a researcher maximizing LLM capabilities for deep editing, like requesting the persona of a professional academic editor. The long prompt for abstract rewriting can be found in Appendix~\ref{sec:other_prompts}, along with the prompt used for title generation.

\section{Results}

\subsection{Word Choice in the Titles}

\begin{figure}[t]
    \begin{subfigure}{\columnwidth}
        \centering
        \includegraphics[width=0.48\columnwidth]{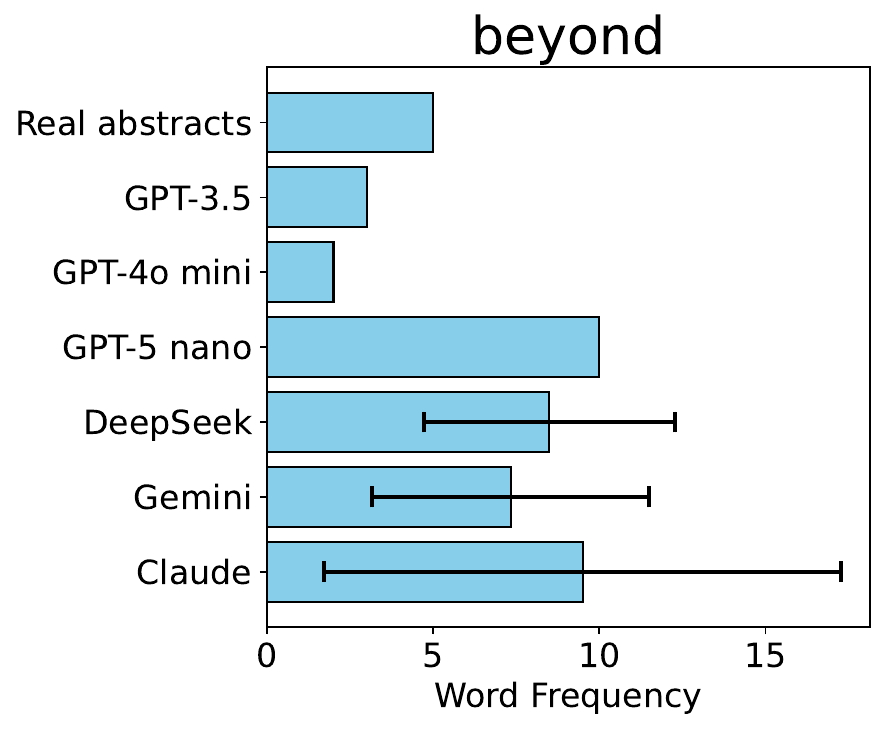}
        \includegraphics[width=0.48\columnwidth]{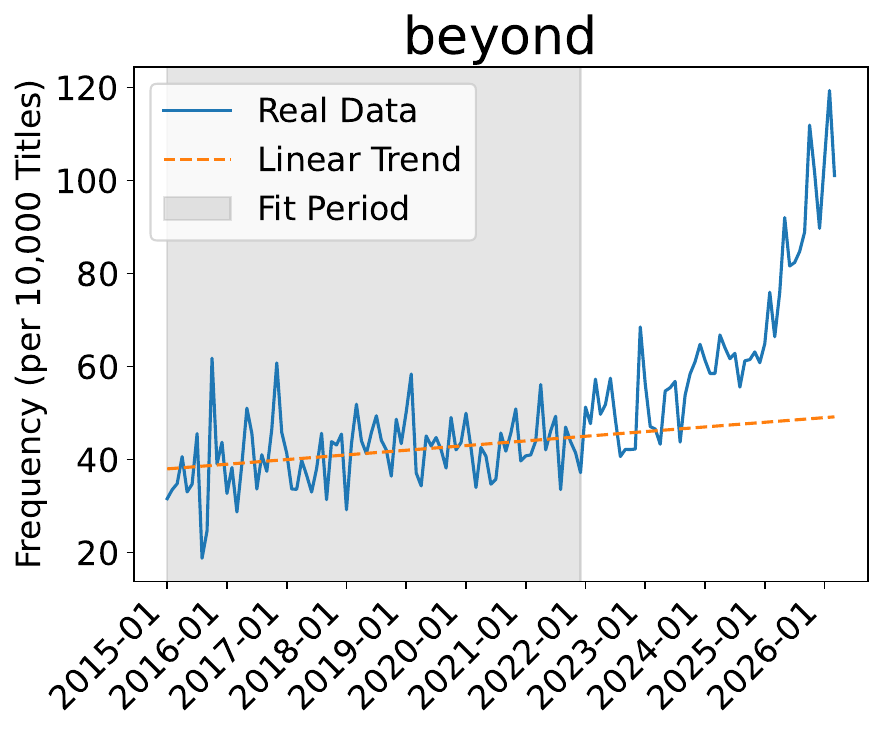}
        \caption{Left: the frequency of the word ``beyond'' in the titles of 2,000 simulated papers. Right: its frequency in actual arXiv titles (across all categories).}
        \label{beyond}
    \end{subfigure}
    \begin{subfigure}{\columnwidth}
        \centering
        \includegraphics[width=0.96\columnwidth]{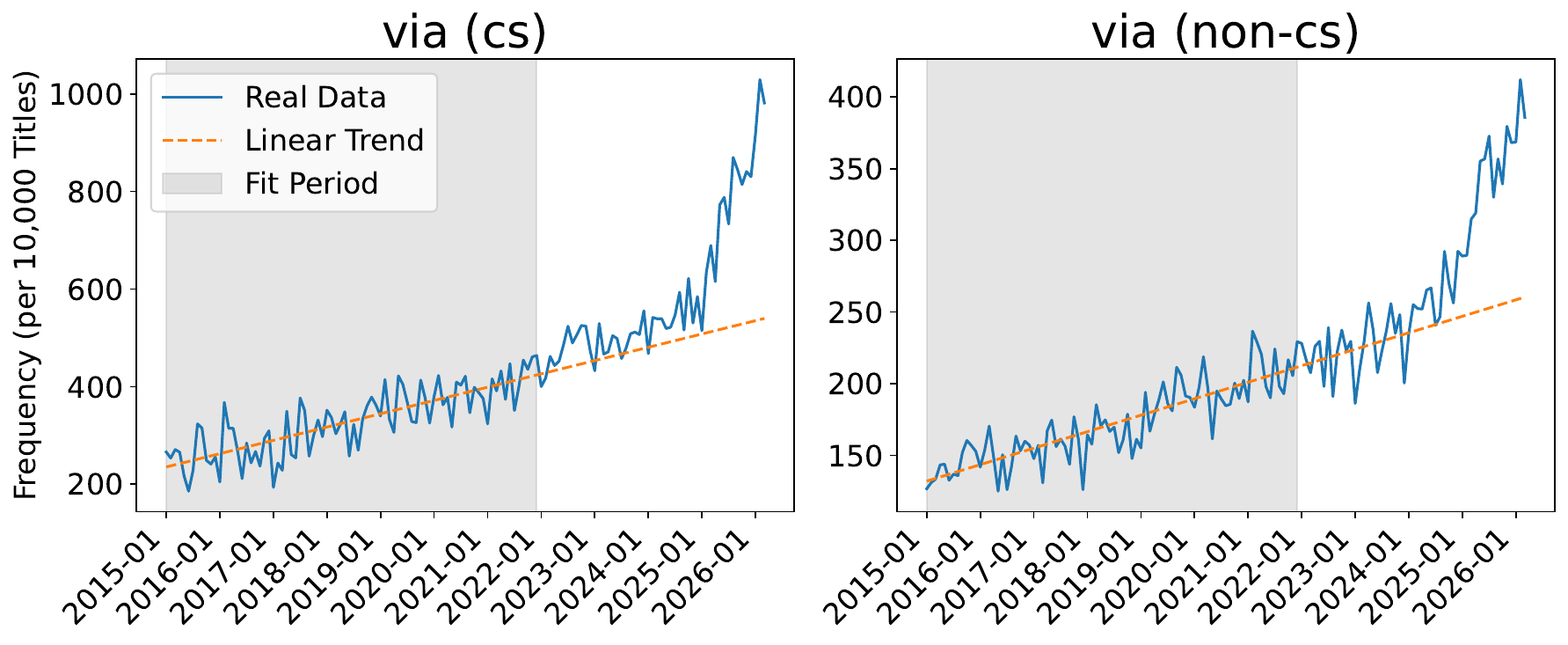}
        \caption{Temporal change in the frequency of the word ``via'' in actual arXiv paper titles for CS and non-CS categories.}
        \label{trend_via_title_cs_non_cs}
    \end{subfigure}
    \caption{Examples of word frequencies in paper titles.}
    \label{trend_llm_title}
    \vspace{-1em}
\end{figure}

LLMs are asked to generate titles based on real abstracts. The word distributions of these titles also differ from those written by humans, which is not surprising. For example, as illustrated in Figures~\ref{via} and \ref{beyond}, newer LLMs such as DeepSeek and GPT-5 tend to favor the words ``via'' and ``beyond'' in paper titles, and their actual frequency only begins to significantly exceed the predicted values starting from 2025. 

We also analyze the differences between CS (computer science) and non-CS papers. As shown in Figure~\ref{trend_via_title_cs_non_cs}, the frequency of ``via'' in titles has increased noticeably in both categories. Given the relatively small number of words in titles, we focus primarily on the analysis of abstracts.

\subsection{Commonly Used Words}

\begin{figure}[t]
    \centering
    \includegraphics[width=0.95\columnwidth]{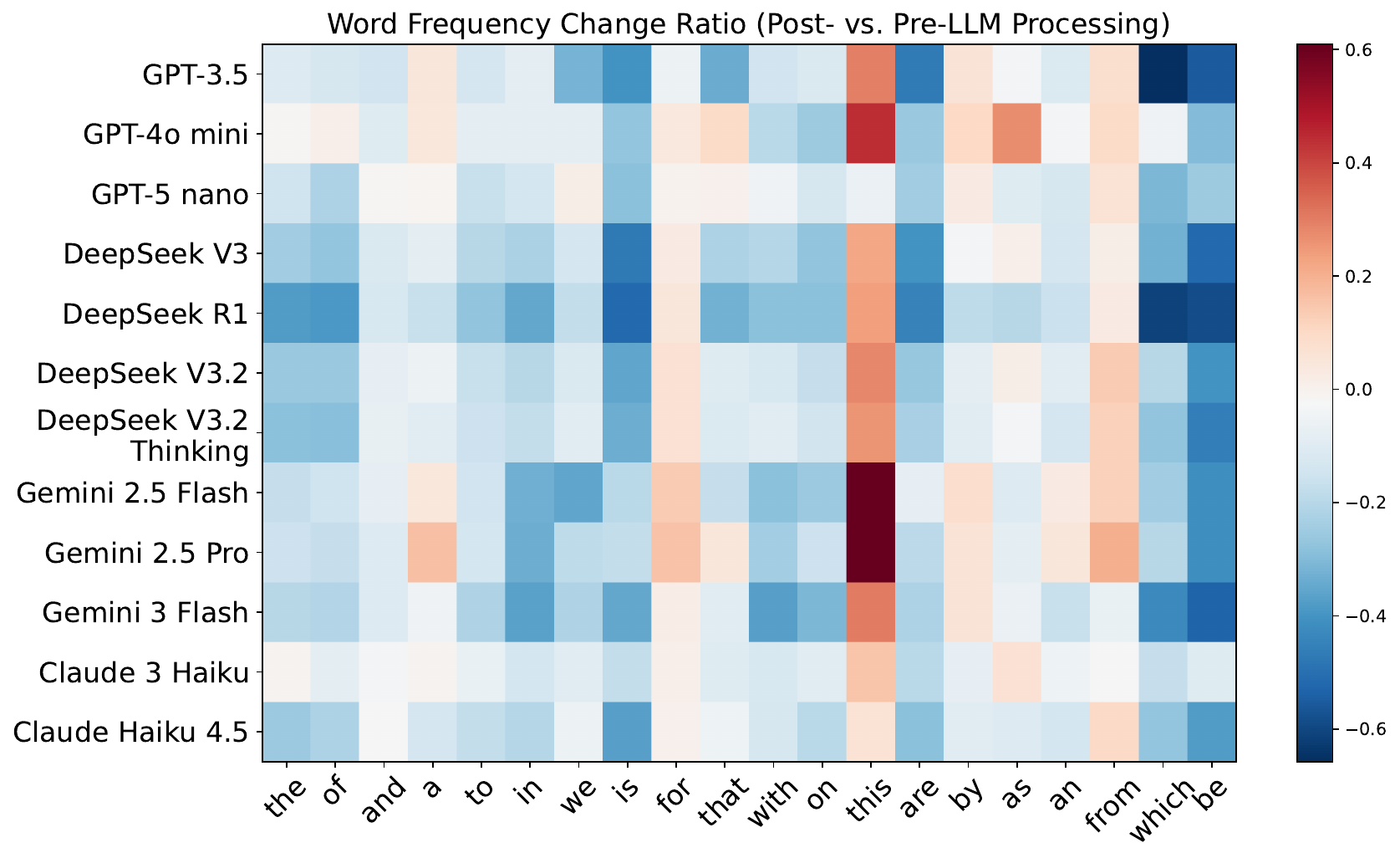}
    \caption{Frequency change ratios of the 20 most frequent words in 2,000 abstracts using multiple LLMs and the shorter prompt.}
    \label{simulation_freq_change_ratio}
    \vspace{-1em}
\end{figure}

\begin{figure}[t]
    \centering
    \includegraphics[width=0.95\columnwidth]{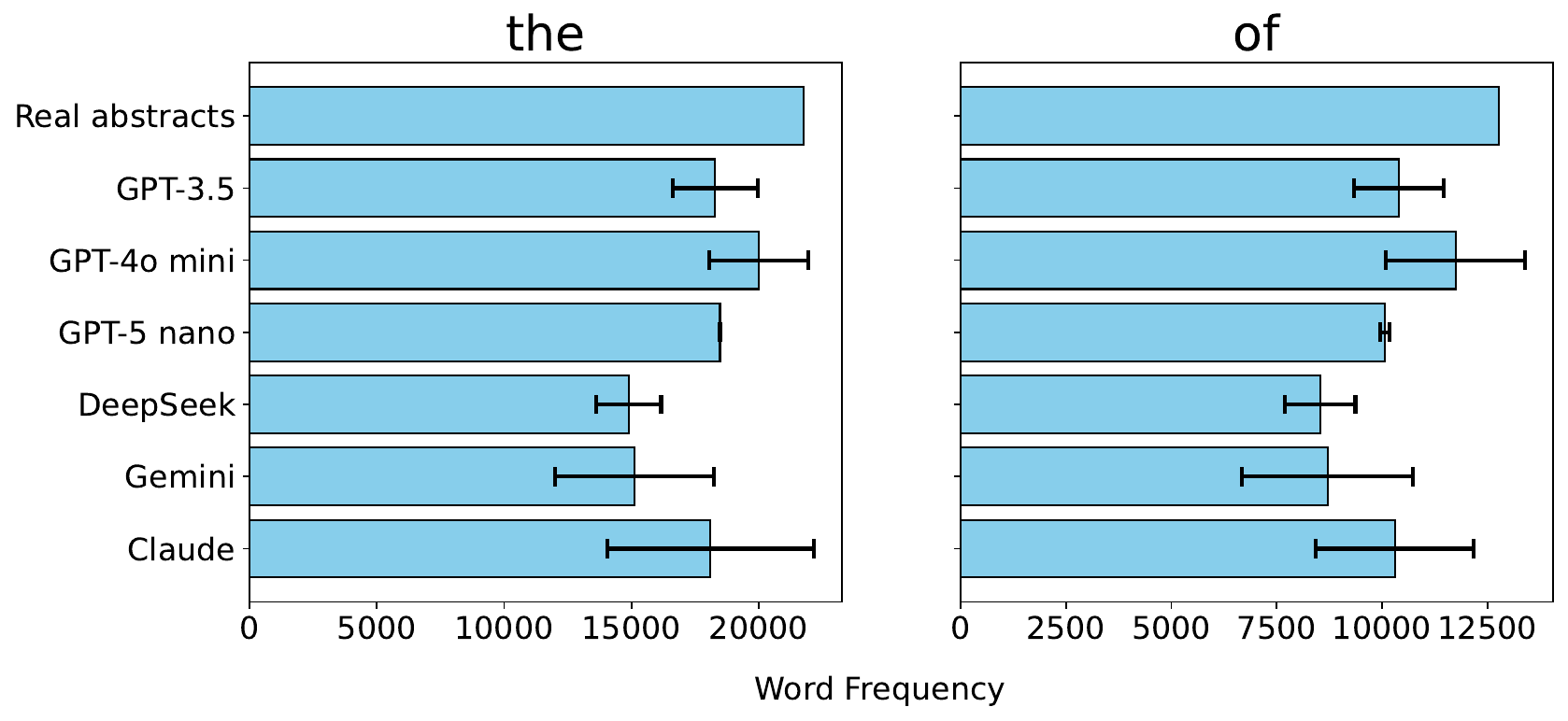}
    \caption{Word frequencies in 2,000 abstracts or in the corresponding LLM-processed content. Error bars denote the standard deviation of word frequencies across outputs produced by different LLMs and/or prompts.}
    \label{llm_the_of}
    \vspace{-1em}
\end{figure}

\begin{figure}[t]
    \begin{subfigure}{\columnwidth}
        \centering
        \includegraphics[width=0.95\columnwidth]{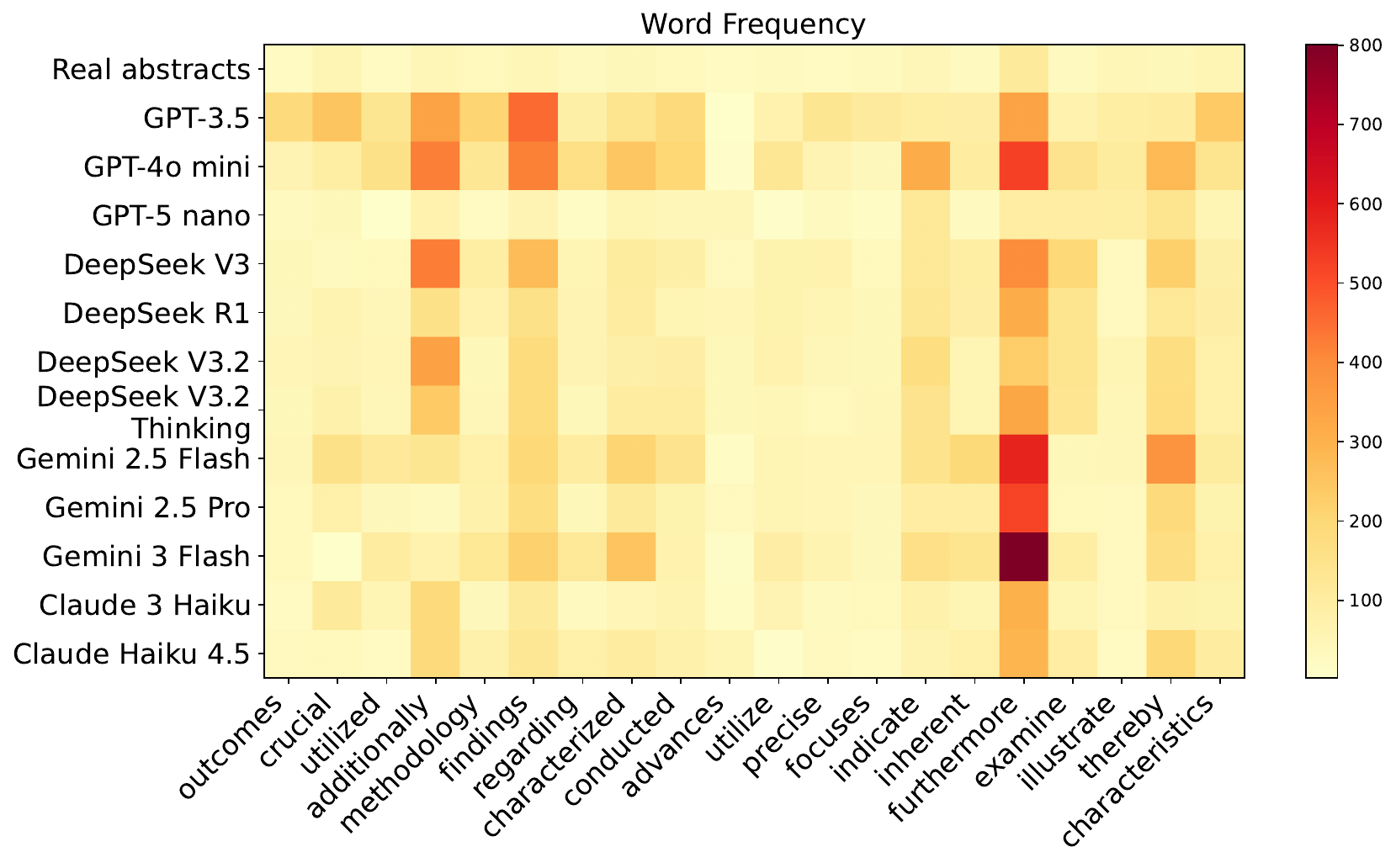}
        \caption{The 20 words with the highest change coefficients that LLMs prefer to use.}
        \label{simulation_more_freq_top_change}
    \end{subfigure}
    \begin{subfigure}{\columnwidth}
        \centering
        \includegraphics[width=0.95\columnwidth]{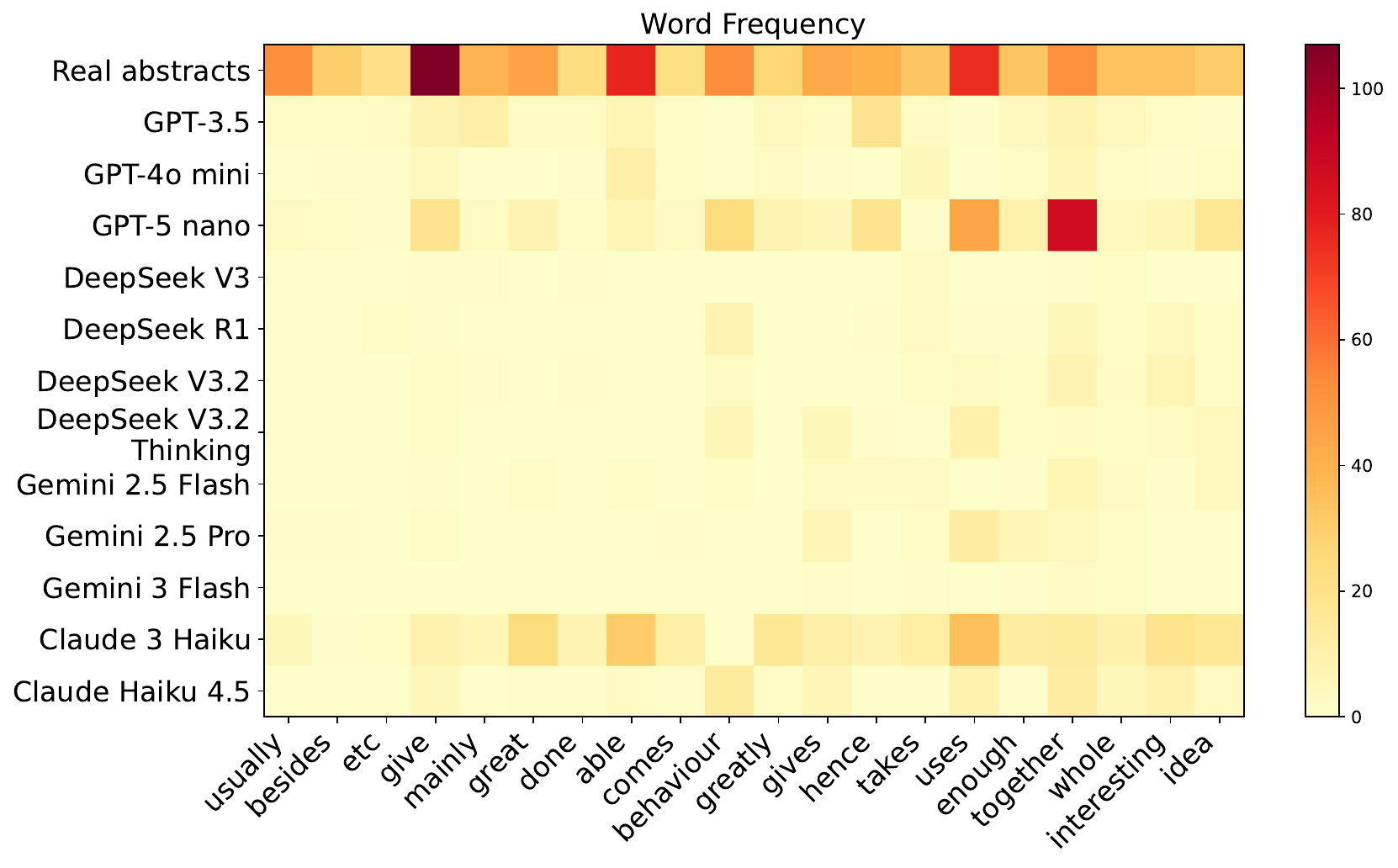}
        \caption{The 20 words with the highest change coefficients that humans prefer to use.}
        \label{simulation_less_freq_top_change}
    \end{subfigure}
    \caption{The frequencies of some words that appeared at least 20 times in the 2,000 abstracts used for the simulation.}
    \vspace{-1em}
\end{figure}

LLMs have their own preferences for word usage. Some commonly used words are often regarded as stopwords that carry little informational content, and are therefore removed in natural language processing~\cite{sarica2021stopwords}, but there are differences between LLMs and humans in the usage of many common words. For instance, for the 20 most common words in the abstracts, their word variation frequency defined in Equation~\ref{r_w_m} is illustrated in Figure~\ref{simulation_freq_change_ratio}. 

Similar to previous results, although it is difficult to estimate LLM usage based on individual words, some correlations can still be observed. The earlier Figure~\ref{together_the_of} shows a decline in the frequencies of ``the'' and ``of'', and Figure~\ref{llm_the_of} demonstrates that most LLMs may indeed avoid using these two words.

To better present word preferences across LLMs, we plot the frequency of the 20 words most favored and the frequency of the 20 words least favored by LLMs, based on the coefficient of variation defined in Equation~\ref{cv}, in Figures~\ref{simulation_more_freq_top_change} and \ref{simulation_less_freq_top_change}.  Thus, in theory, the changes in the frequencies of these common words can also be used to estimate the influence of LLMs on academic publications.

\subsection{Changes and Evolution of LLMs}

LLMs are also continuously evolving, and different LLMs have distinct word preferences. The word ``together'' in Figure~\ref{together_the_of} serves as a clear example. Words previously considered characteristic of ChatGPT, like ``\textbf{delve}'' and ``\textbf{intricate}'', are  abandoned by newer LLMs, as seen in Figure~\ref{llm_delve_intricate} in the Appendix. 

There are also more examples of the change in word usage preferences, including even some more common words. For instance, as illustrated in Figure~\ref{and_this_furthermore_thereby} in the appendix, the frequency of ``\textbf{and}'' has recently exceeded the predicted upward trend, whereas the frequency of ``\textbf{this}'' first decreased and then increased. These shifts may reflect differences in word preference between some recent models such as GPT-5 nano and earlier models, as shown in Figure~\ref{simulation_freq_change_ratio}.

In addition, Figure~\ref{simulation_more_freq_top_change} indicates that, except for GPT-5 Nano, other models show a strong preference for the word ``\textbf{furthermore}''. The frequency of this term rises sharply but returns to a level close to its previous growth trajectory recently, as illustrated in Figure~\ref{and_this_furthermore_thereby} of the appendix. This pattern could also indicate that GPT-5 Nano or models like it are more widely used in academic writing. 

The changes in word frequency in abstracts discussed above are observed across the entire arXiv corpus. Similar patterns also emerge when CS and non-CS papers are examined separately. For example, Figure~\ref{the_of_cs_non_cs} in the appendix shows the frequency changes of ``the'', ``of'', ``delve'', and ``intricate'' in these two groups. Interactive results for more words and categories are available on the website\footnote{\url{https://llm-impact.github.io/word-usage-arxiv-abstract/}}.

The changes and evolution of LLMs also require us to estimate and monitor their impact with a dynamic perspective. The differences among various LLMs offer valuable insights for estimating the usage and influence of different models. 

\subsection{Text Similarity Analysis}

\begin{figure}[!t]
    \begin{subfigure}{\columnwidth}
        \centering
        \includegraphics[width=0.98\columnwidth]{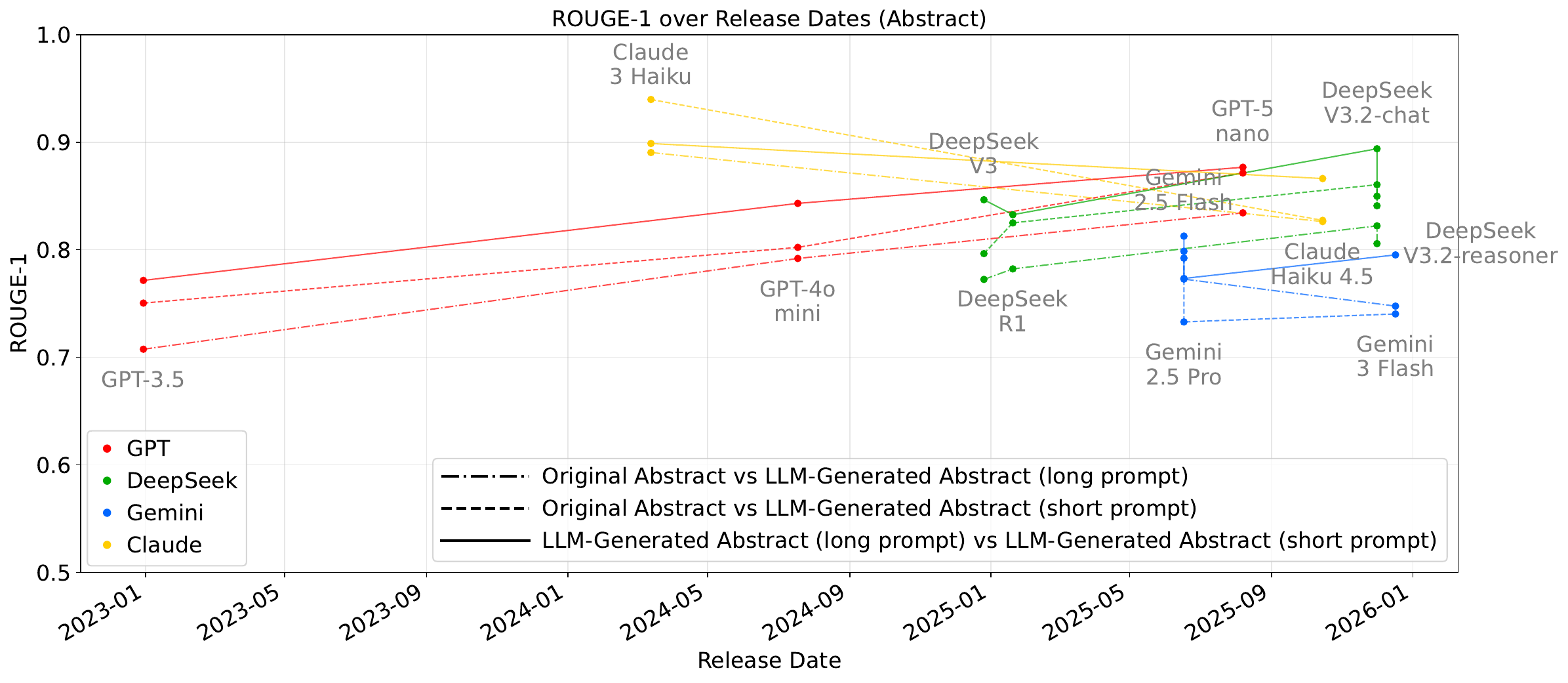}
        \includegraphics[width=0.98\columnwidth]{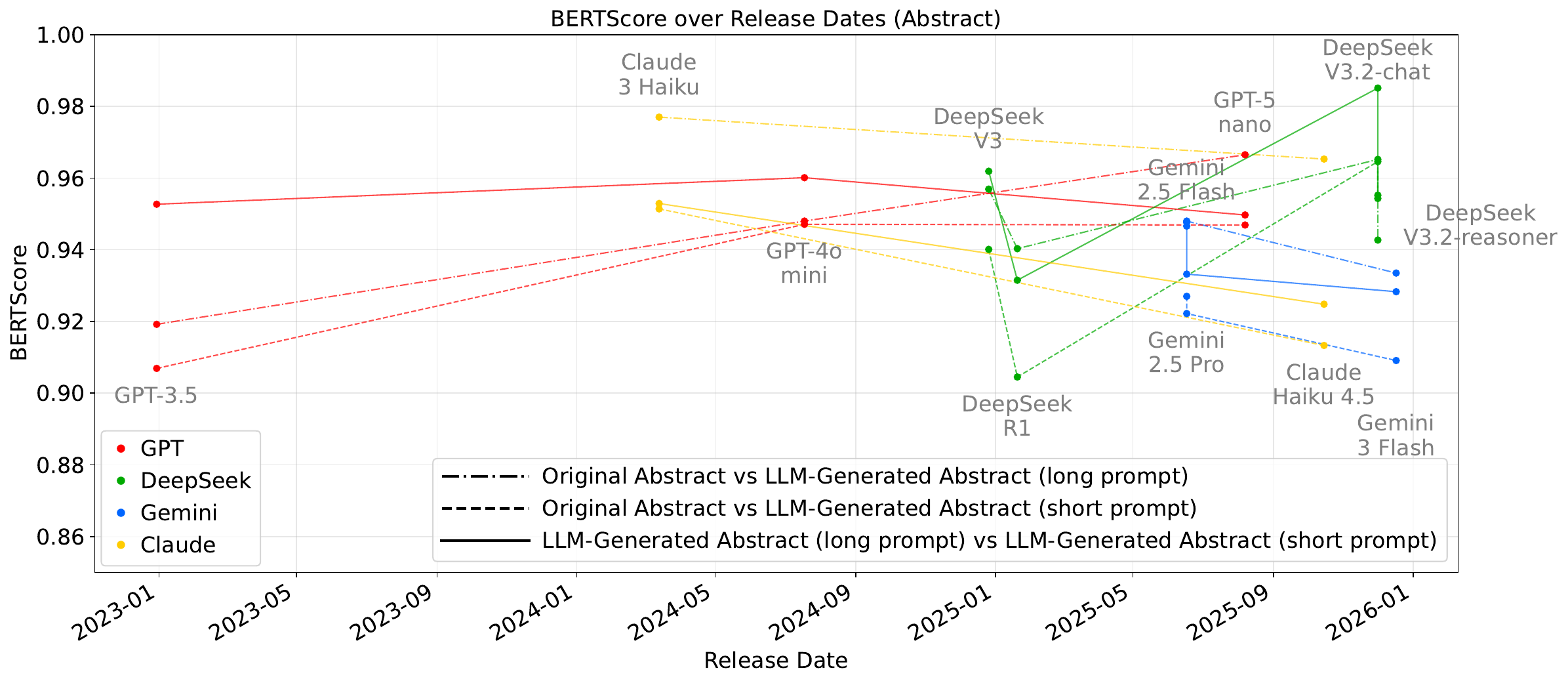}
        \caption{Similarity between the processed texts of the two prompts and their similarity to the real abstracts.}
        \label{abstract_metrics_grid_v1}
    \end{subfigure}
        \begin{subfigure}{\columnwidth}
        \centering
        \includegraphics[width=0.98\columnwidth]{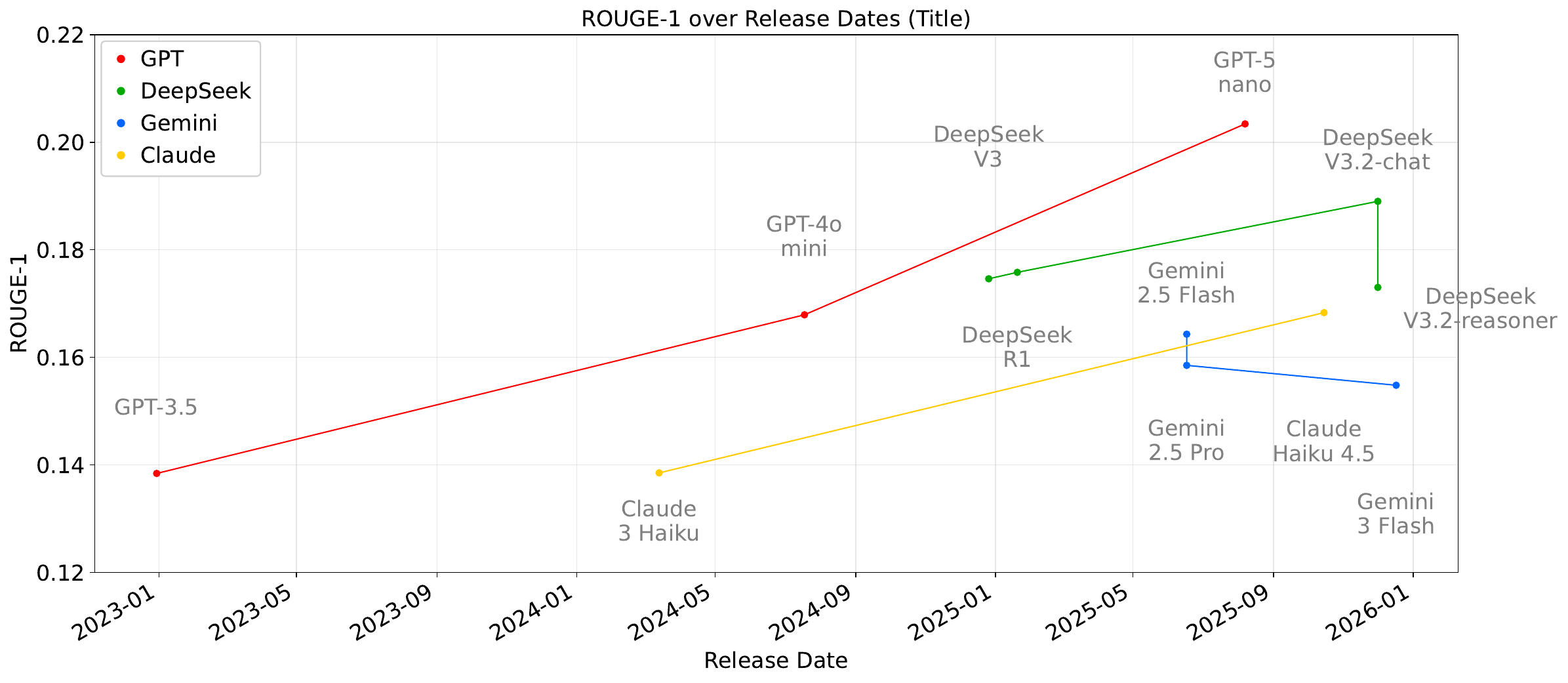}
        \includegraphics[width=0.98\columnwidth]{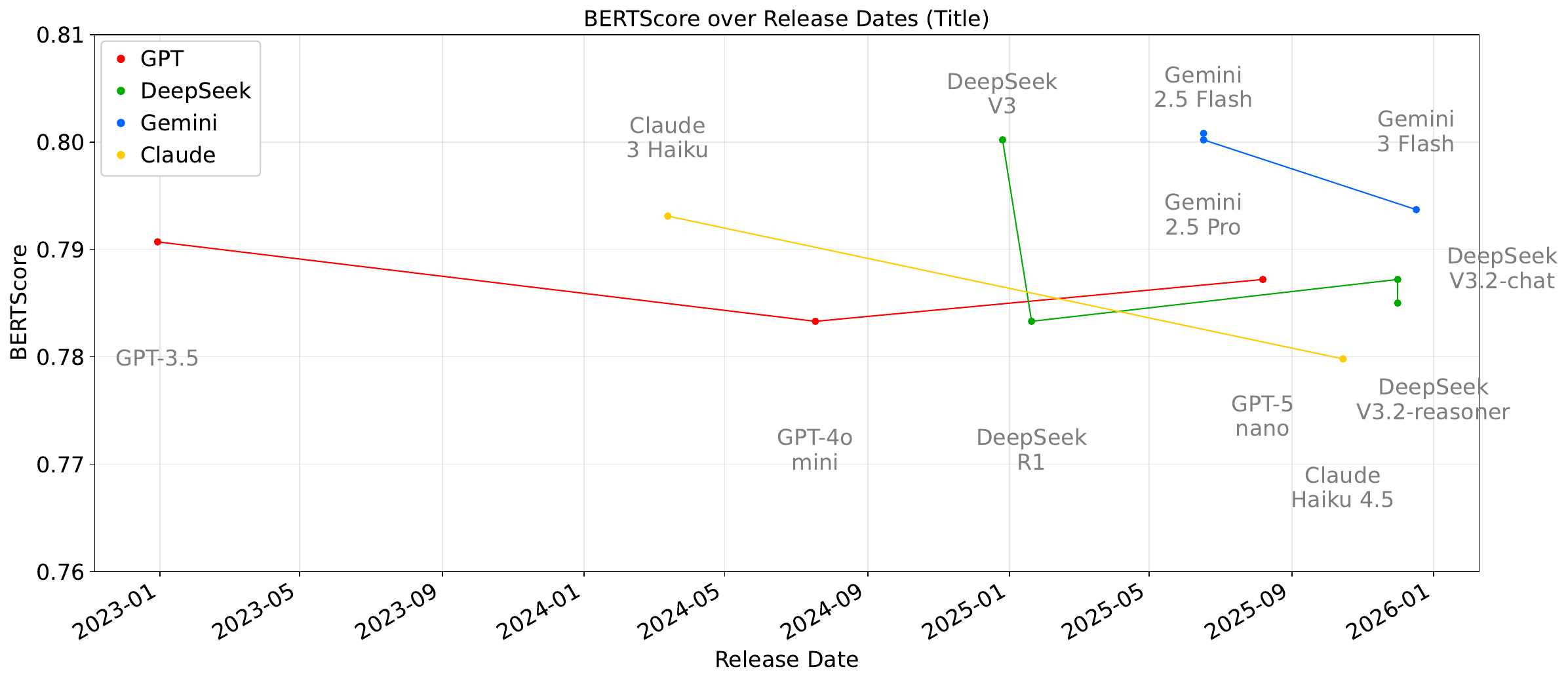}
        \caption{Similarity between the titles generated by LLMs and the real titles.}
        \label{title_metrics_grid_v1}
    \end{subfigure}
    \caption{Comparison Results of Text Similarity. The x-axis represents the model release dates.}
    \label{metrics_grid_v1}
    \vspace{-1em}
\end{figure}

Apart from the differences in word usage frequency, we aim to examine it from a broader perspective of text similarity (e.g., ROUGE-1 and BERTScore). The comparison between the original abstracts and titles and the content processed or generated by LLMs, as well as the comparison of the same model's outputs across different prompts, are shown in Figure~\ref{metrics_grid_v1}. The more comprehensive comparison results can be found in Figure~\ref{metrics_grid_v2} of the Appendix.

Figures~\ref{abstract_metrics_grid_v1} and \ref{abstract_metrics_grid_v2} indicate that the abstracts processed by the two prompts are close to each other, whereas they show a larger difference from the original ones. Claude Haiku 3 stands as an exception, as the abstracts resulting from the shorter prompt are more similar to the original abstracts. In general, the results from the two prompts of the new LLMs are gradually aligning more closely with the original abstracts under ROUGE metrics. However, semantically, this trend is not as evident.

Similar conclusions can be drawn from the title generation in Figures~\ref{title_metrics_grid_v1} and \ref{title_metrics_grid_v2}: while the new LLMs are closer to the true titles according to ROUGE, this is not reflected in the semantic results from BERTScore.

The analysis in this part further highlights the differences between various LLMs, as well as the changes occurring within models of the same series. It is important to note that higher similarity does not necessarily mean stronger model performance, which may simply indicate that the model has made fewer modifications to the output text.

\subsection{Classification}

\begin{table}[t]
\centering
\caption{Classification results. Prompt: long, Classifier: BERT, Model Type: old. The black squares indicate the source of the detected text, and the accuracy shown is the highest value obtained across the parameter set.}
\label{tab:long_BERT_old}

\setlength{\tabcolsep}{2.5pt}
{
\begin{tabular}{lccccccc}
\toprule
\textbf{GPT}      & $\blacksquare$ & $\blacksquare$ & $\blacksquare$ & $\blacksquare$ & \textcolor{gray!30}{$\cdot$} & \textcolor{gray!30}{$\cdot$} & \textcolor{gray!30}{$\cdot$} \\
\textbf{DeepSeek} & $\blacksquare$ & $\blacksquare$ & \textcolor{gray!30}{$\cdot$} & \textcolor{gray!30}{$\cdot$} & $\blacksquare$ & $\blacksquare$ & \textcolor{gray!30}{$\cdot$} \\
\textbf{Gemini}   & $\blacksquare$ & \textcolor{gray!30}{$\cdot$} & $\blacksquare$ & \textcolor{gray!30}{$\cdot$} & $\blacksquare$ & \textcolor{gray!30}{$\cdot$} & $\blacksquare$ \\
\textbf{Claude}   & $\blacksquare$ & \textcolor{gray!30}{$\cdot$} & \textcolor{gray!30}{$\cdot$} & $\blacksquare$ & \textcolor{gray!30}{$\cdot$} & $\blacksquare$ & $\blacksquare$ \\
\midrule
\textbf{Acc (\%)} & 78.6 & 90.1 & 90.1 & \textbf{95.3} & 76.9 & 93.4 & 91.8 \\
\bottomrule
\end{tabular}
}
\end{table}

\begin{table}[t]
\centering
\caption{Classification results. Prompt: long, Classifier: BERT, Model Type: new.}
\label{tab:long_BERT_new}

\setlength{\tabcolsep}{2.5pt}

\begin{tabular}{lccccccc}
\toprule
\textbf{GPT} & $\blacksquare$ & $\blacksquare$ & $\blacksquare$ & $\blacksquare$ & \textcolor{gray!30}{$\cdot$} & \textcolor{gray!30}{$\cdot$} & \textcolor{gray!30}{$\cdot$} \\
\textbf{DeepSeek} & $\blacksquare$ & $\blacksquare$ & \textcolor{gray!30}{$\cdot$} & \textcolor{gray!30}{$\cdot$} & $\blacksquare$ & $\blacksquare$ & \textcolor{gray!30}{$\cdot$} \\
\textbf{Gemini} & $\blacksquare$ & \textcolor{gray!30}{$\cdot$} & $\blacksquare$ & \textcolor{gray!30}{$\cdot$} & $\blacksquare$ & \textcolor{gray!30}{$\cdot$} & $\blacksquare$ \\
\textbf{Claude} & $\blacksquare$ & \textcolor{gray!30}{$\cdot$} & \textcolor{gray!30}{$\cdot$} & $\blacksquare$ & \textcolor{gray!30}{$\cdot$} & $\blacksquare$ & $\blacksquare$ \\
\midrule
\textbf{Acc (\%)} & 63.0 & 83.0 & \textbf{90.4} & 84.9 & 83.8 & 70.9 & 88.2 \\
\bottomrule
\end{tabular}
\end{table}

\begin{figure}[!t]
    \centering
    \includegraphics[width=0.95\columnwidth]{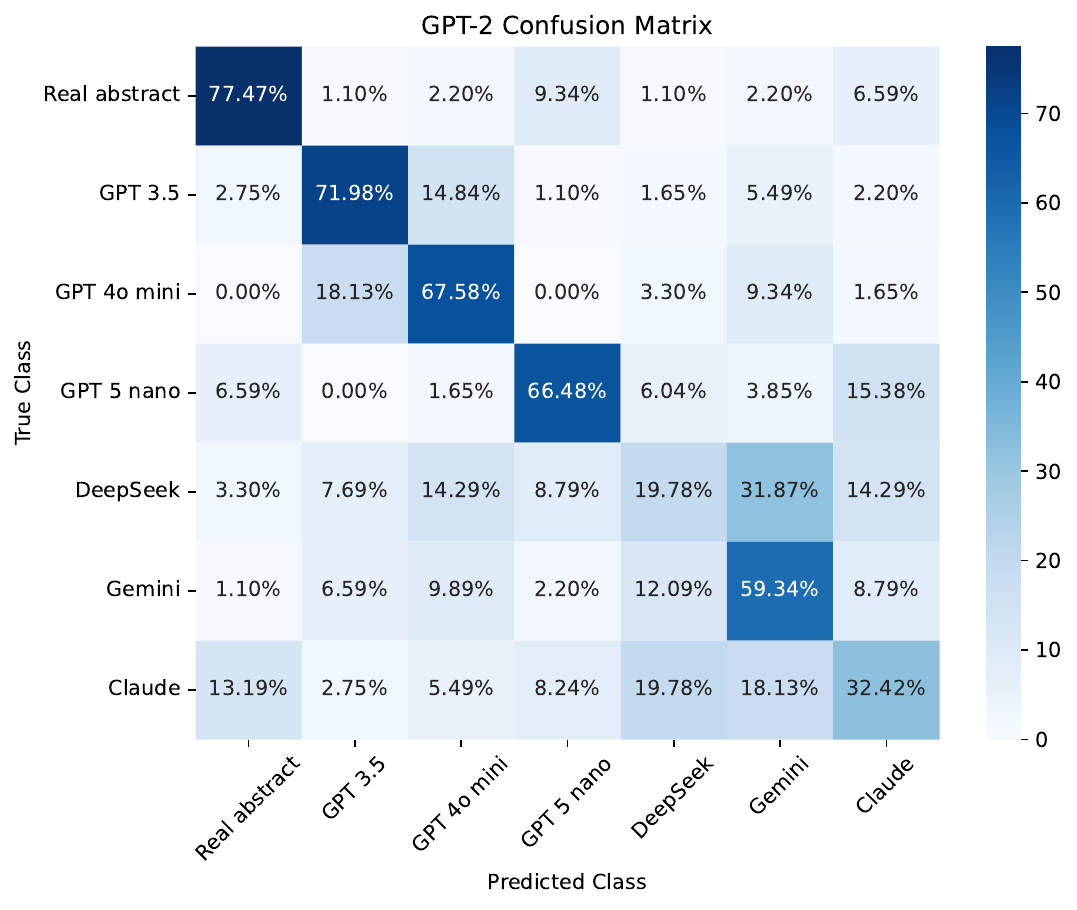}
    \includegraphics[width=0.95\columnwidth]{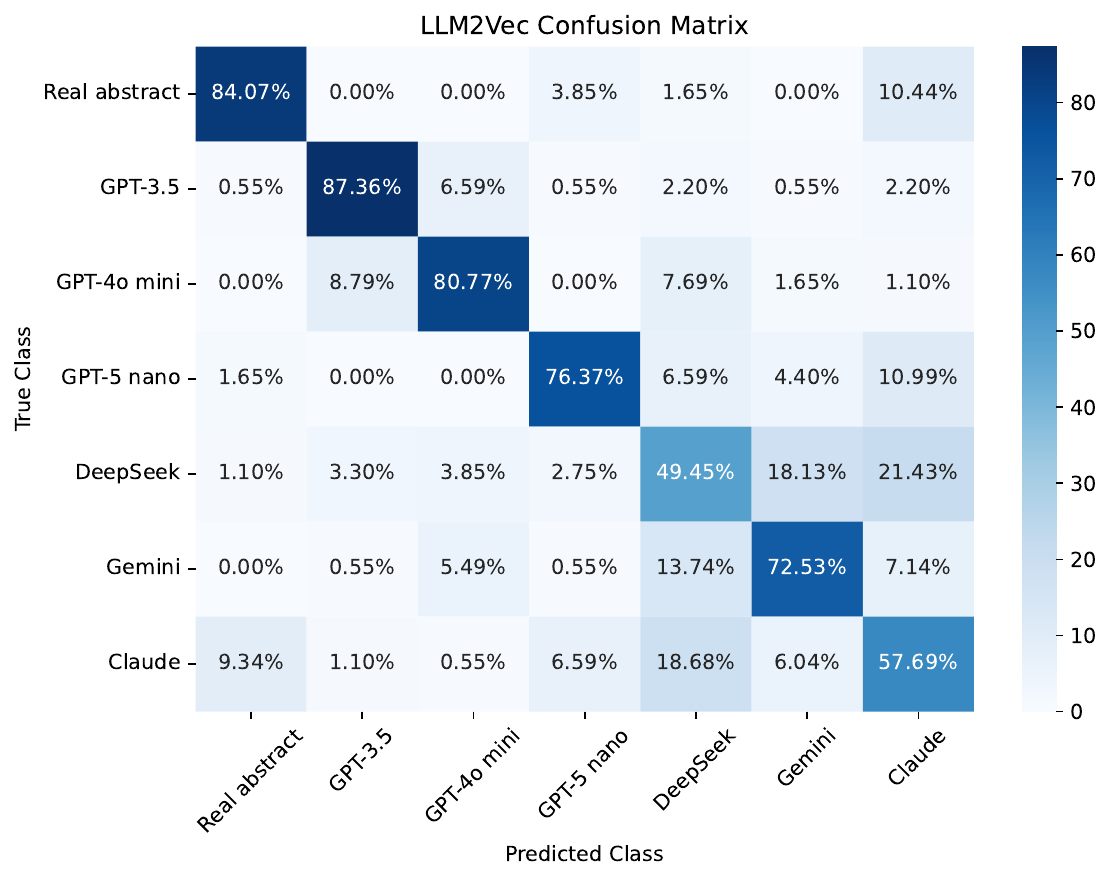}
    \caption{Confusion matrix of classification results from the detectors based on GPT-2 and LLM2Vec. In the training and test data, GPT-3.5, GPT-4o mini, and GPT-5-nano are mixtures of two prompts, while DeepSeek, Gemini, and Claude are mixtures of multiple versions and two prompts. After mixing, each class contains 2,000 abstracts.}
    \label{GPT-2_confusion_matrix}
\end{figure}

We consider two scenarios in classification tasks: distinguishing between texts generated by different LLMs, and a multi-class classification that includes texts written by humans. The method and code we use here are based on the rare open-source work that involves classifying texts from different LLMs~\cite{sun2025idiosyncrasies}.

In the first scenario, we consider binary and four-class classification. An earlier version (GPT-3.5, DeepSeek V3, Gemini 2.5 Flash, Claude 3 Haiku) and the most recent version (GPT-5 nano, DeepSeek V3.2, Gemini 3 Flash, Claude Haiku 4.5) from each of the four model families (GPT, DeepSeek, Gemini, and Claude) are chosen for comparison.

Tables~\ref{tab:long_BERT_old} and \ref{tab:long_BERT_new} present the classification results of the BERT-based classifier for texts generated with long prompt, while Table~\ref{tab:long_GPT-2_old} to Table~\ref{tab:short_T5_new} in the Appendix provide more comparison results. The values in the table represent the results with the highest accuracy in the test set with different parameters. In general, the results indicate that the abstracts polished by different LLMs show some differences, with binary classification accuracy reaching around 80\%-90\% and four-class classification accuracy at about 60\%. In the old models, the long prompt generally achieves higher classification accuracy than the short prompt. In other words, the differences between different LLMs are narrowing, which may indicate a homogenizing effect.

For the second scenario, we directly perform a seven-class or thirteen-class classification, including texts written by humans. The results of the seven categories shown in Figure~\ref{GPT-2_confusion_matrix} and Figure~\ref{confusion_matrix_gpt_t5} of the appendix refer to three GPT models, along with DeepSeek, Gemini, and Claude (with mixed outputs from different models in each series), as well as human-written texts. In Figures~\ref{fig:confmat-bert} to \ref{fig:confmat-llm2vec} of the appendix, the thirteen categories correspond to the classification of each model separately. The table~\ref{tab:combined_classification} in the Appendix summarizes some of the results.

It can be observed from these results that the accuracy declines with the increase in categories. While the accuracy of classifying human-written texts and GPT-processed texts is quite high, it's also fairly common for human-written text to be mistakenly identified as LLM-generated. In real-world scenarios, the number of prompts is far greater than two, and there are more models involved than those in our experiment.

\subsection{Impact Estimation}

\begin{figure}[t]
    \begin{subfigure}{\columnwidth}
        \centering
        \includegraphics[width=0.95\columnwidth]{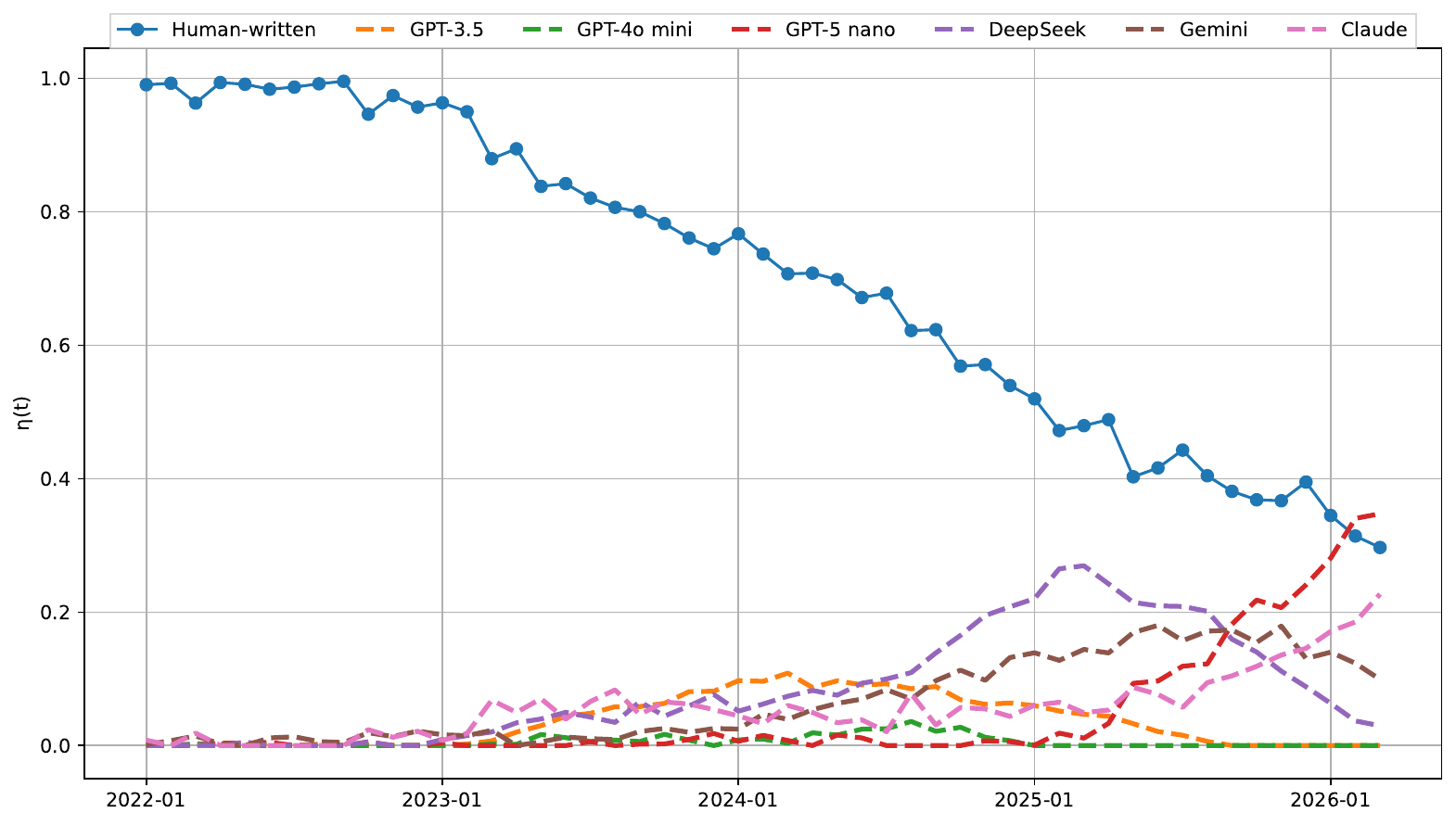}
        \caption{The 2,741 words used for estimation are drawn from the 10,000 most frequent words in the Google Books Ngram dataset that occur at least 10 times in the 2,000 preprocessed abstracts used for simulation.}
        \label{eta_estimation_10_0_10000}
    \end{subfigure}
        \begin{subfigure}{\columnwidth}
        \centering
        \includegraphics[width=0.95\columnwidth]{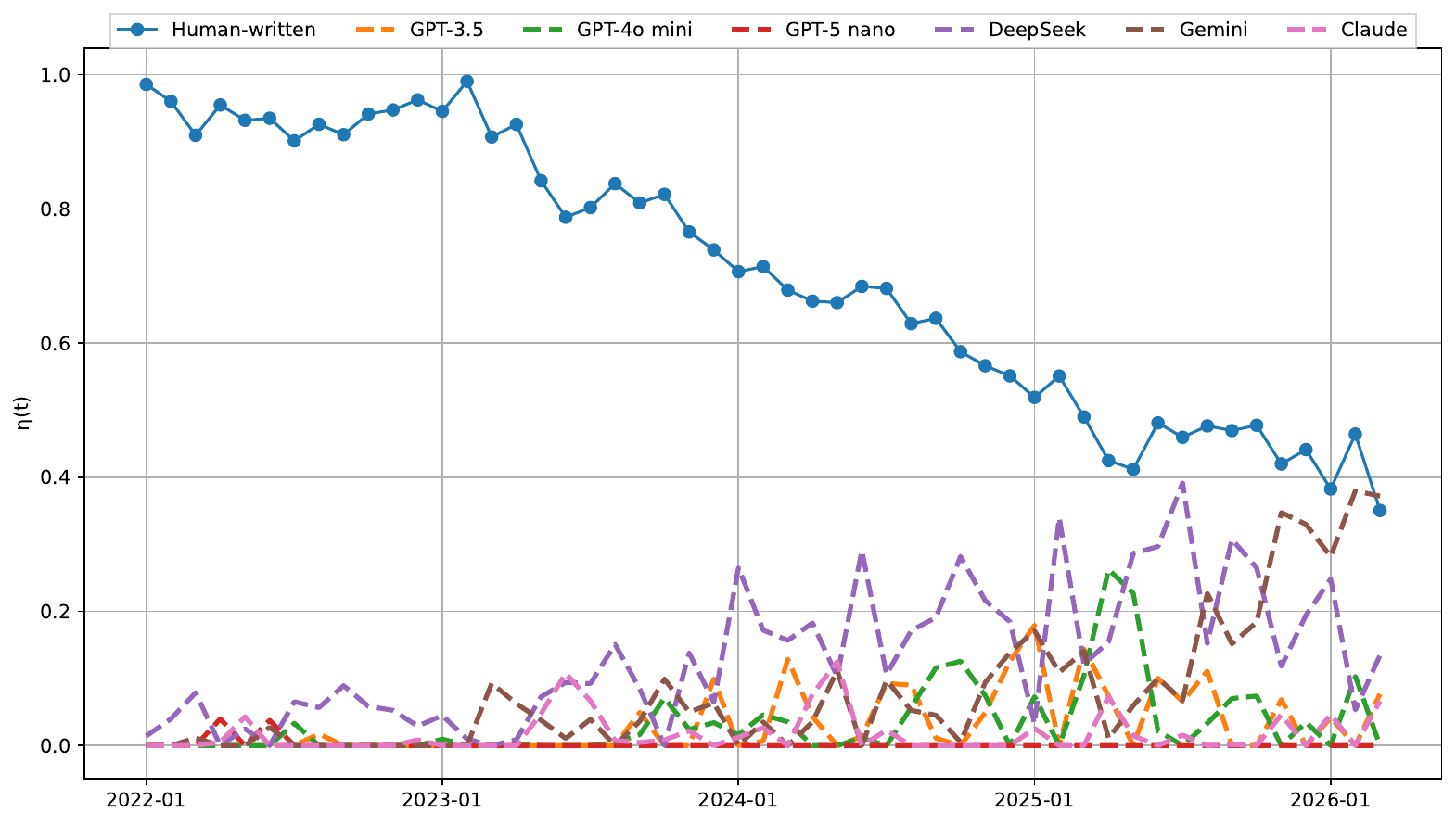}
        \caption{The 126 words used for estimation are drawn from NLTK's English stopwords in the Google Books Ngram dataset that occur in the 2,000 cleaned abstracts used for simulation.}
        \label{eta_estimation_stopwords}
    \end{subfigure}
    \caption{Estimation of LLM impact in arXiv abstracts.}
    \label{eta_estimation_results}
\end{figure}

Given that real abstracts have about 20\% or even more chance of being misclassified as LLM-generated with the classifier described above, we use the previously described word frequency-based estimation, which is easier to interpret. Based on Equation~\ref{eq_eta_estimation}, the estimated impact of LLMs varies depending on the different words and parameters. For example, we can set
\begin{equation}
    \eta_{m,p}(t) \in [0, 1], \quad \forall m \in \mathcal{M^*}, \, p \in \mathcal{P^*}
\end{equation}

To demonstrate the effectiveness of the simple method, we used SLSQP (Sequential Least Squares Programming) to solve this equation, with the initial value of $\eta_{m,p}(t)$ uniformly distributed. Figure~\ref{eta_estimation_results} presents the estimates derived from the models in Table~\ref{tab:llm_release_dates} and the prompts in Section~\ref{prompts}. 

The outputs of various LLMs show some similarities~\cite{jiang2025artificial}, which could reduce the accuracy of predictions. The results in Figure~\ref{eta_estimation_10_0_10000} are based on 2,741 words that are relatively common in both the Google Books Ngram dataset and arXiv abstracts, while the results in Figure~\ref{eta_estimation_stopwords} are derived from the stopwords in NLTK that appear at least once in the 2,000 arXiv abstracts. We can also observe in Figure~\ref{eta_estimation_10_0_10000} that the proportion of text similar to the GPT-3.5 style first increases and then decreases. The estimates based on stopwords in Figure~\ref{eta_estimation_stopwords} show larger fluctuations, likely due to the low differentiation of these words across different models, even though they differ from human-written text.

These results are generally consistent with our expectations, such as the LLM impact estimates being close to zero before October 2022. Figure~\ref{eta_estimation_results_supplementary} in the appendix also presents results estimated using different combinations of words. The optimal choice of term combinations can be determined in multiple ways, for instance constructing a validation dataset with known ground truth by mixing real abstracts with LLM-generated ones~\citep{geng2024chatgpt}. Alternatively, one can examine which estimates yield the smallest misclassification values before the emergence of ChatGPT or other models.

\section{Related Work}

\paragraph{LLM-Generated Text Detection.} Researchers have proposed many detectors for identifying text generated by LLMs, with a variety of approaches and techniques~\cite{wu2025survey}. However, in real-world scenarios, such as academic writing, human modifications to LLM outputs can indeed affect the effectiveness of detection~\cite{russell2025people}. Considering the variety of models and use cases, it may be an impossible task to accurately estimate the numerical usage of LLMs~\cite{geng2025detectability}. %

\paragraph{LLM Fingerprint.} Some LLMs each have their own linguistic fingerprints, such as in vocabulary and lexical diversity, the frequency of certain lexical and morphosyntactic features~\cite{reviriego2024playing,mcgovern2025your}. These features can be used to trace their origins~\cite{nikolic2025model} as well as to detect LLM-generated text~\cite{antoun2024text,sun2025idiosyncrasies} and check API calls~\cite{gao2024model}. Similar studies have also appeared in other languages, although some LLMs may be quite similar and difficult to distinguish~\cite{zaitsu2025stylometry}.

\paragraph{LLMs for Scientific Writing.} Many people are discussing the use of LLMs in scientific research, and this has become an unstoppable trend~\citep{eger2025transforming}. The disclosure of AI usage may impact how people perceive the article~\citep{hazra2025accepted}. Despite the widespread use of AI in academic writing, there are few papers that disclose its usage~\cite{he2025academic}. The scope of LLMs' use in scientific research goes beyond just writing, such as research fields~\citep{hao2026artificial}, citations~\cite{algaba2025deep}, and more.

\paragraph{Impact Estimation} The influence of LLMs on academic publications is diverse~\cite{kusumegi2025scientific}. In addition to the previously mentioned works, some papers attempt to quantify these impacts, such as through the use of words~\cite{liang2025quantifying} and different linguistic shifts~\cite{bao2025examining,lin2025chatgpt}. 

\section{Discussion}

The focus of this paper is on the impact of LLMs rather than on whether the text was generated by them. Classifiers are likely to perform well in binary classification, but their accuracy significantly decreases in multi-class tasks. In real-world scenarios, however, the latter situation is more common. 

Although the method of analyzing word frequency may sound simple, this intuitive approach proves to be quite effective in analyzing the impact of LLMs. Focusing on more common words may provide better estimates, and our simple method can fill the gap left by complex classifiers. Our estimates can also be improved, for example, by considering more models and prompts. 

In the era of LLMs, the way people use words is continually evolving~\cite{geng2025human,mak2025style}. The effect of LLMs on text goes \textbf{beyond} academic publications. As similar impacts are likely to increase in the future, the methods and findings presented in this paper could serve as important references for subsequent research. The language style of LLMs, including their preference for certain words, requires further exploration~\cite{juzek2025does}. 

The opportunities these tools provide differ for authors from various regions (e.g., Eastern vs. Western)~\cite{khan2025gets}, though our paper does not explore this aspect. The modification of content and information is more troubling than any change in wording~\citep{mohamed2025llm,abdulhai2026llms}. The generation of content may require more caution, such as hallucinated references~\cite{sakai2026hallucitation}. 

\section{Conclusion}

We systematically analyze similarities and differences between arXiv abstracts and titles processed by different LLMs. Compared to human-written texts, LLM-generated outputs exhibit distinct stylistic characteristics, with additional variation observed across different models. 

While complex classifiers can be employed for detection, the application of such black-box methods in real-world scenarios raises concerns. Nevertheless, it remains possible to estimate the influence of LLMs by examining the differences between their outputs and human-written text. Our approach provides a more direct and intuitive illustration of the impact of LLMs on academic writing. The results show that LLMs, especially the GPT series, have a significant impact on the writing in arXiv abstracts, but other models likely have considerable usage as well. 

Considering both similarity comparisons and detector outcomes, model outputs may indeed become progressively more human-like, raising more challenges for AI-generated text detection. Furthermore, humans could be subtly shaped by machines. Consequently, approaches for tracking and evaluating their impact will need to be continuously refined as LLMs advance.

\section*{Acknowledgments}

This work benefited from funding from the French State, managed by the Agence Nationale de la Recherche, under the France 2030 program (grant reference ANR-23-IACL-0008). This research also received support from the ENS-PSL BeYs Chair in Data Science and Cybersecurity.

\bibliography{paper}
\bibliographystyle{icml2026}

\appendix
\clearpage

\section{Simulation Details}

\subsection{Other prompts}
\paragraph{Long prompt for abstract rewriting:}
\label{sec:other_prompts}
    \begin{quote}
    \small\ttfamily
    Role:\\
    Act as a professional academic editor and reviewer for top-tier scientific journals (e.g., Nature, IEEE Transactions, ACM). You have expertise in technical writing, grammar, and scientific logic.
    
    Task:\\
    Refine and polish the specific text provided below. Your goal is to improve clarity, coherence, and academic tone while strictly maintaining the original technical meaning.
    
    Constraints \& Guidelines:\\
    1. Grammar \& Syntax: Correct all grammatical, spelling, and punctuation errors.\\
    2. Academic Tone: Use formal, objective, and precise language. Avoid colloquialisms, contractions, or overly flowery language.\\
    3. Clarity \& Flow: Improve sentence structure to enhance readability. Break down overly complex sentences if necessary, but ensure the logical flow remains smooth.\\
    4. Vocabulary: Replace weak or repetitive words with more precise academic vocabulary suited for a technical context.\\
    5. Preservation:\\
    \hspace*{1em}- Do NOT change the core scientific meaning or data.\\
    \hspace*{1em}- Do NOT alter specific variable names, LaTeX formulas, citations (e.g., [1]), or technical terminology unless they are clearly incorrect.\\
    6. Conciseness: Remove redundancy and fluff.
    
    Output Format:\\
    Please only output the polished text in JSON format. Only one version. No explanations. Only plain text.
    
    EXAMPLE JSON OUTPUT:\\
    \{"text": "Your polished version of the provided text goes here."\}
    \end{quote}

\paragraph{Prompts for title generation:}
    \begin{quote}
    \small\ttfamily
    Read the following abstract of a research paper. Generate a representative title that accurately reflects the main contribution of the work.\\

    Constraint: You must output ONLY a valid JSON object. Do not include any introductory or concluding text.\\

    Format: \{"title": "Your generated title here"\}
    \end{quote}

\subsection{API Usage}
\label{api_usage}

All polished texts utilized in this study were generated via cloud-based LLM services rather than local execution. To ensure data consistency and mitigate potential disruptions caused by network instability, all LLM responses were acquired using Python-based batch API processing instead of interactive chat interfaces. The APIs were configured to output data in JSON format to strictly exclude extraneous explanatory text. All other settings were retained at their default values.

The specific model APIs employed are listed below:

\paragraph{1. OpenAI}
Models: \texttt{gpt-3.5-turbo}, \texttt{gpt-4o-mini}, \texttt{gpt-5-nano}. \\
Access period: January 13, 2025 -- January 14, 2025.

\paragraph{2. DeepSeek}
Models: \texttt{deepseek-chat}, \texttt{deepseek-reasoner}. \\
Note: DeepSeek provides only the latest model versions on its official website without specific version reference. According to the official documentation, the actual version that was executed during the access period was DeepSeek-V3.2. \\
Access period: January 13, 2025 -- January 15, 2025.

\paragraph{3. Alibaba Cloud}
Models: \texttt{deepseek-v3}, \texttt{deepseek-r1}. \\
Note: These models utilized open-source code provided by DeepSeek and were hosted on servers operated by the Alibaba Cloud internet vendor. \\
Access period: January 13, 2025 -- January 16, 2025.

\paragraph{4. Google}
Models: \texttt{gemini-2.5-flash}, \texttt{gemini-2.5-pro}, \texttt{gemini-3-flash-preview}. \\
Access period: January 13, 2025 -- January 15, 2025.

\paragraph{5. Anthropic}
Models: \texttt{claude-3-haiku-20240307}, \texttt{claude-haiku-4-5-20251001}. \\
Access period: January 21, 2026.

\section{Other Results}

\begin{figure}[t]
    \begin{subfigure}{\columnwidth}
        \centering
        \includegraphics[width=0.95\columnwidth]{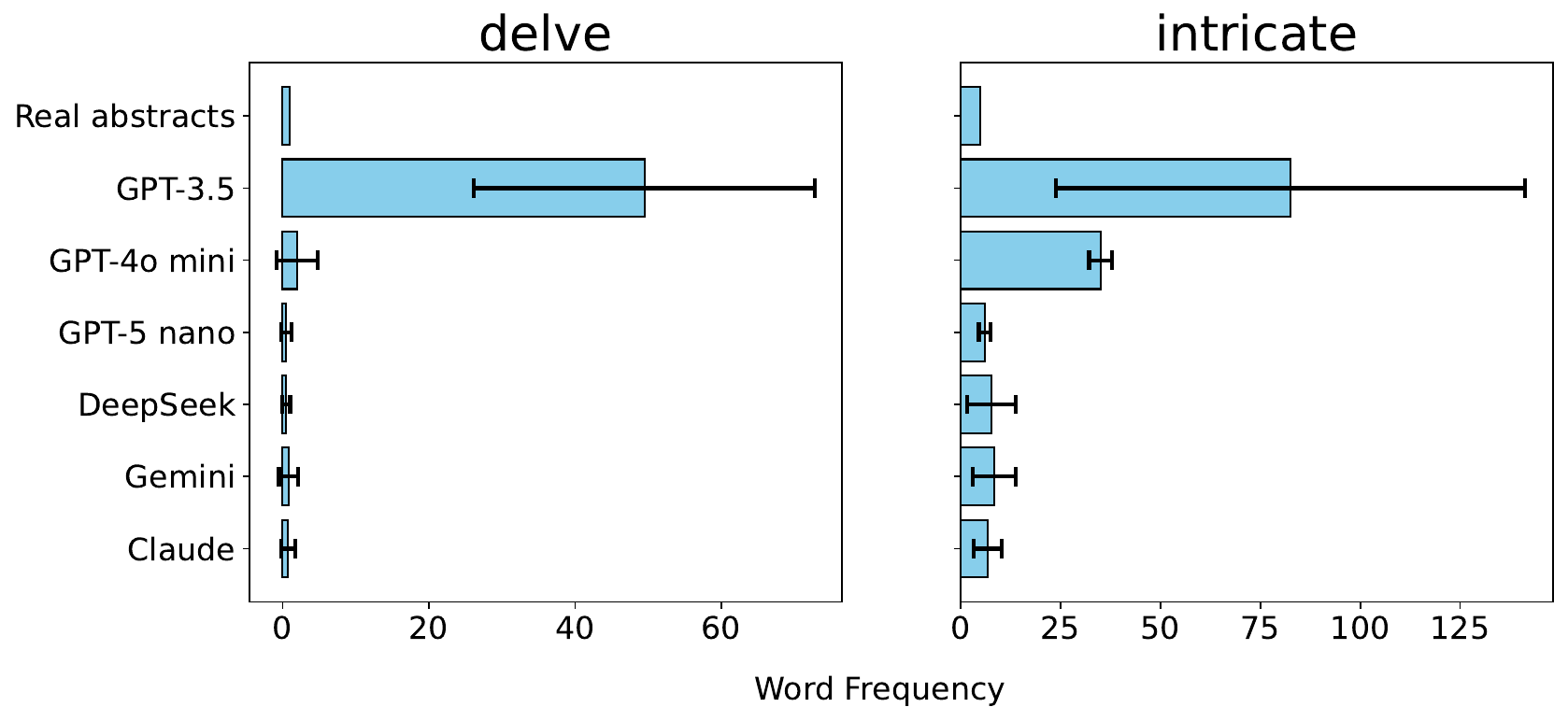}   
        \includegraphics[width=0.95\columnwidth]{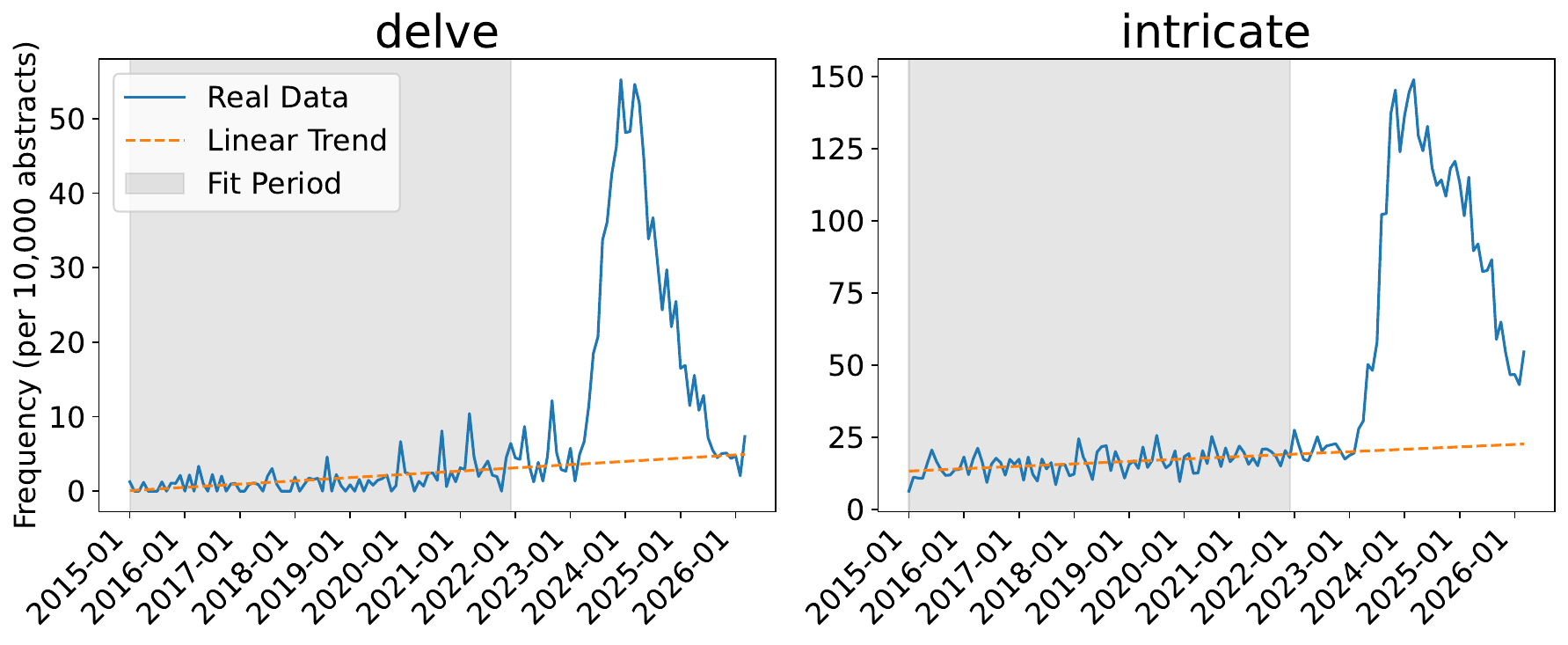}  
        \subcaption{Two words favored by GPT-3.5.}
        \label{llm_delve_intricate}
    \end{subfigure}
    \begin{subfigure}{\columnwidth}
        \centering
        \includegraphics[width=0.95\columnwidth]{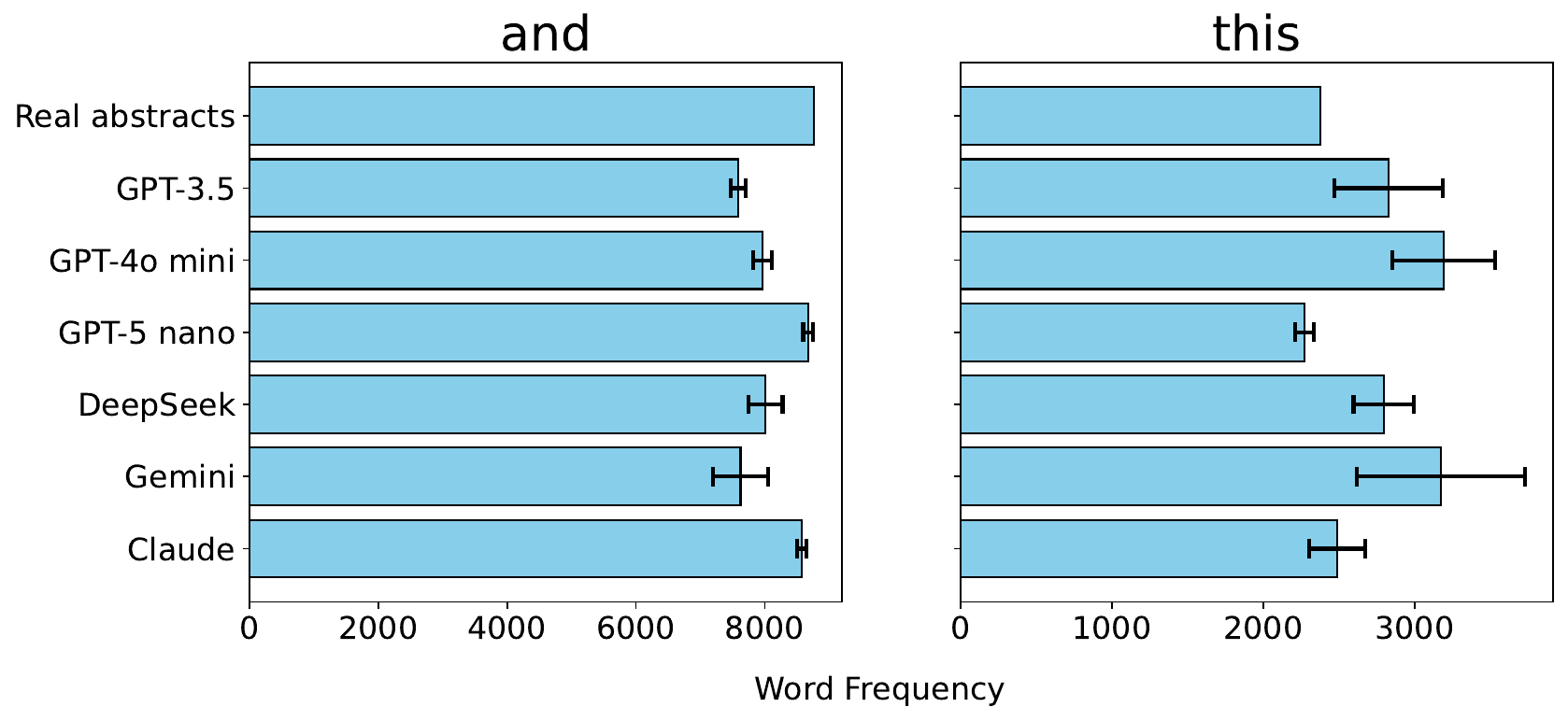}    
        \includegraphics[width=0.95\columnwidth]{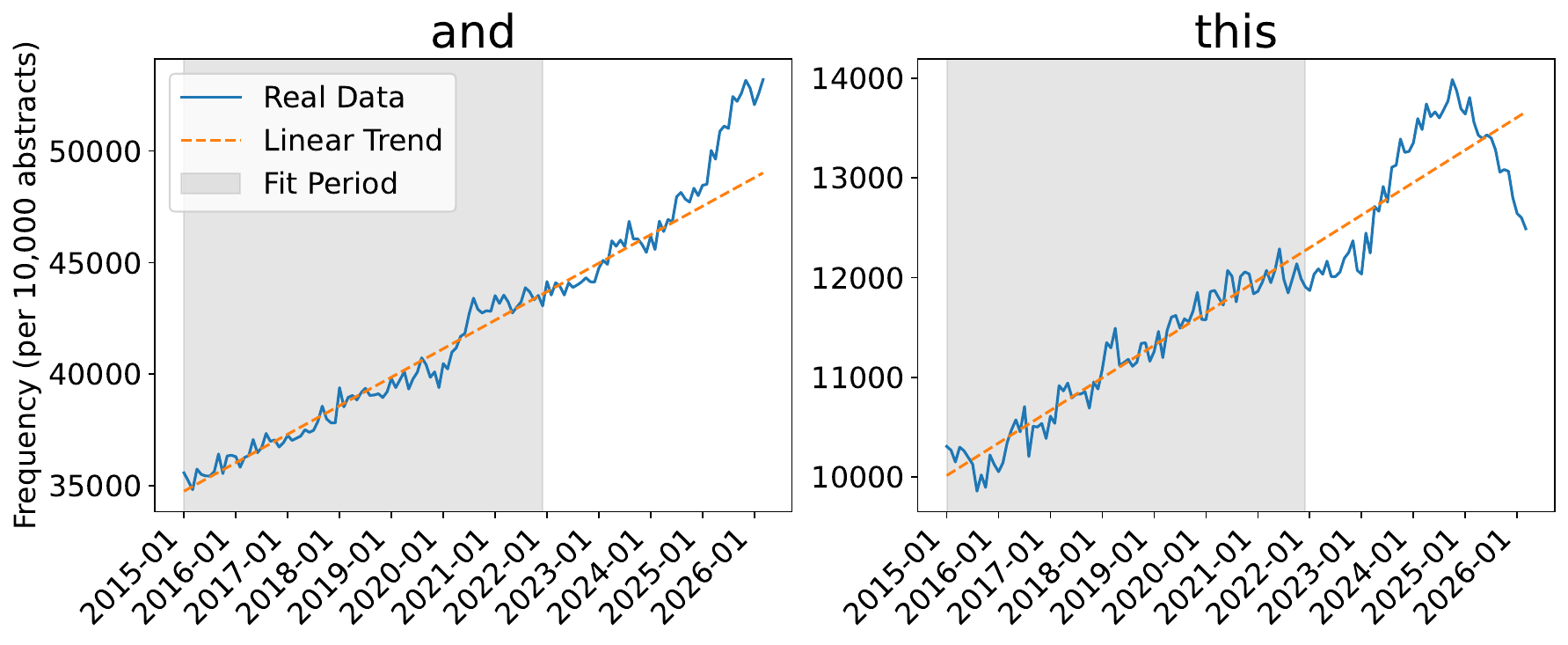}
        \includegraphics[width=0.95\columnwidth]{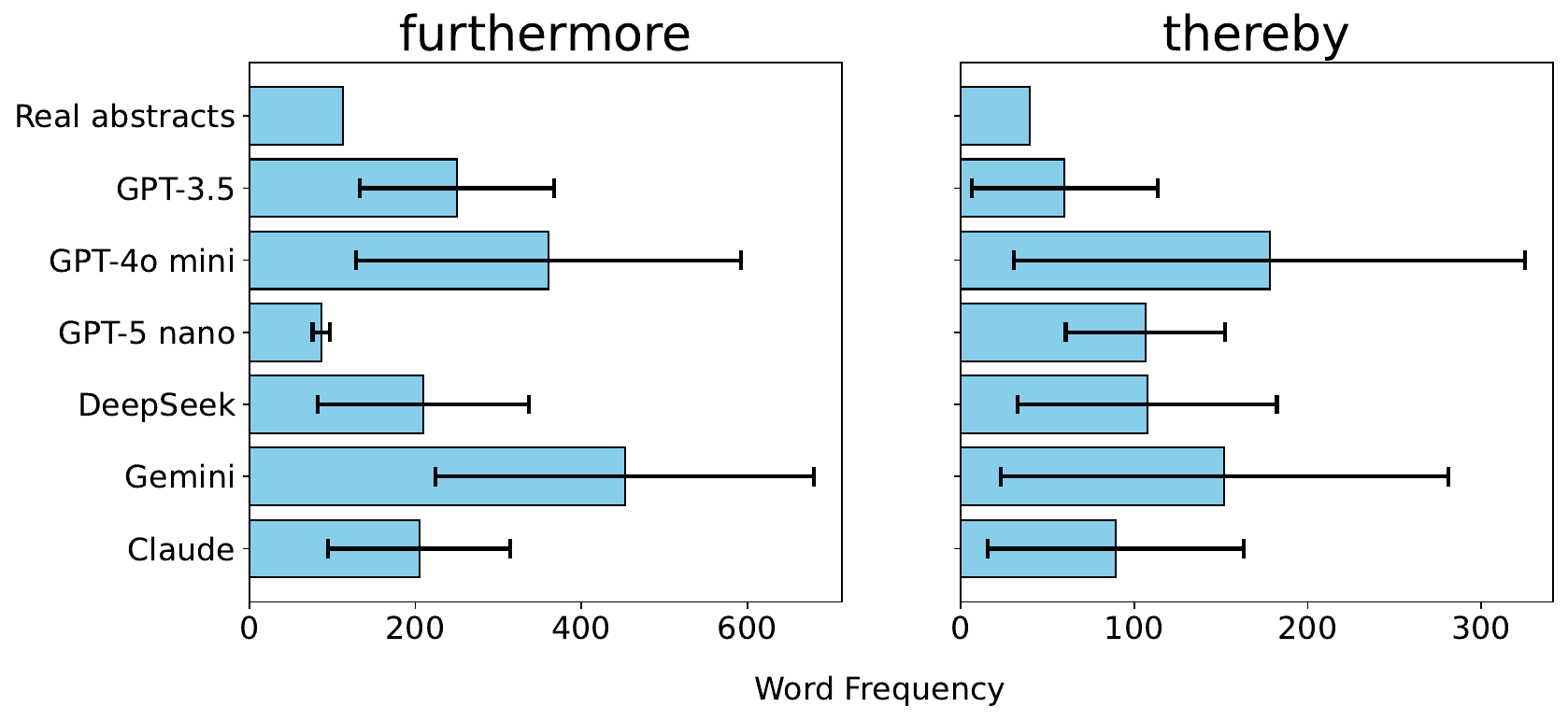}
        \includegraphics[width=0.95\columnwidth]{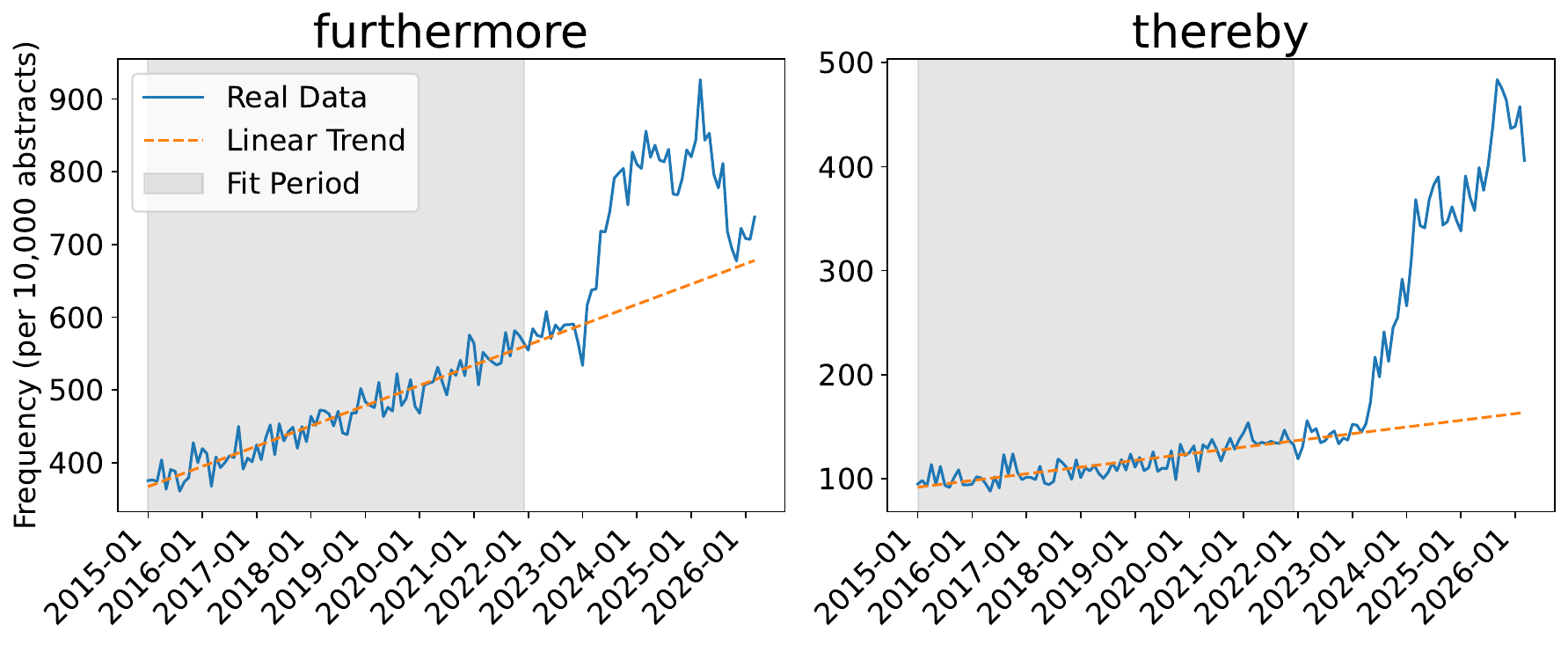}
        \subcaption{Four common words.}
        \label{and_this_furthermore_thereby}
    \end{subfigure}
    \caption{Supplementary comparison results.}
\end{figure}

\begin{figure}[t]
    \begin{subfigure}{\columnwidth}
        \centering
        \includegraphics[width=0.95\columnwidth]{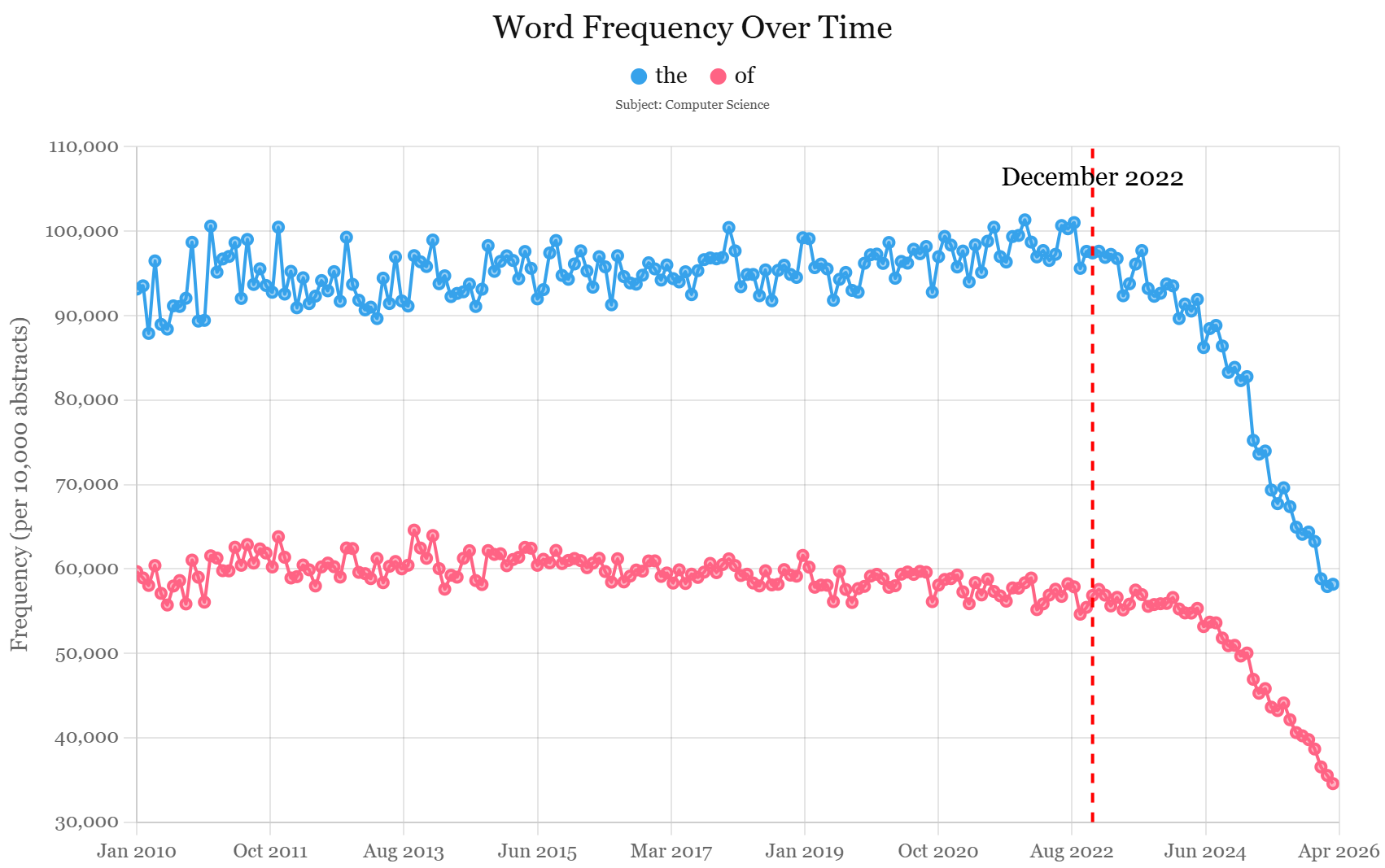}
        \includegraphics[width=0.95\columnwidth]{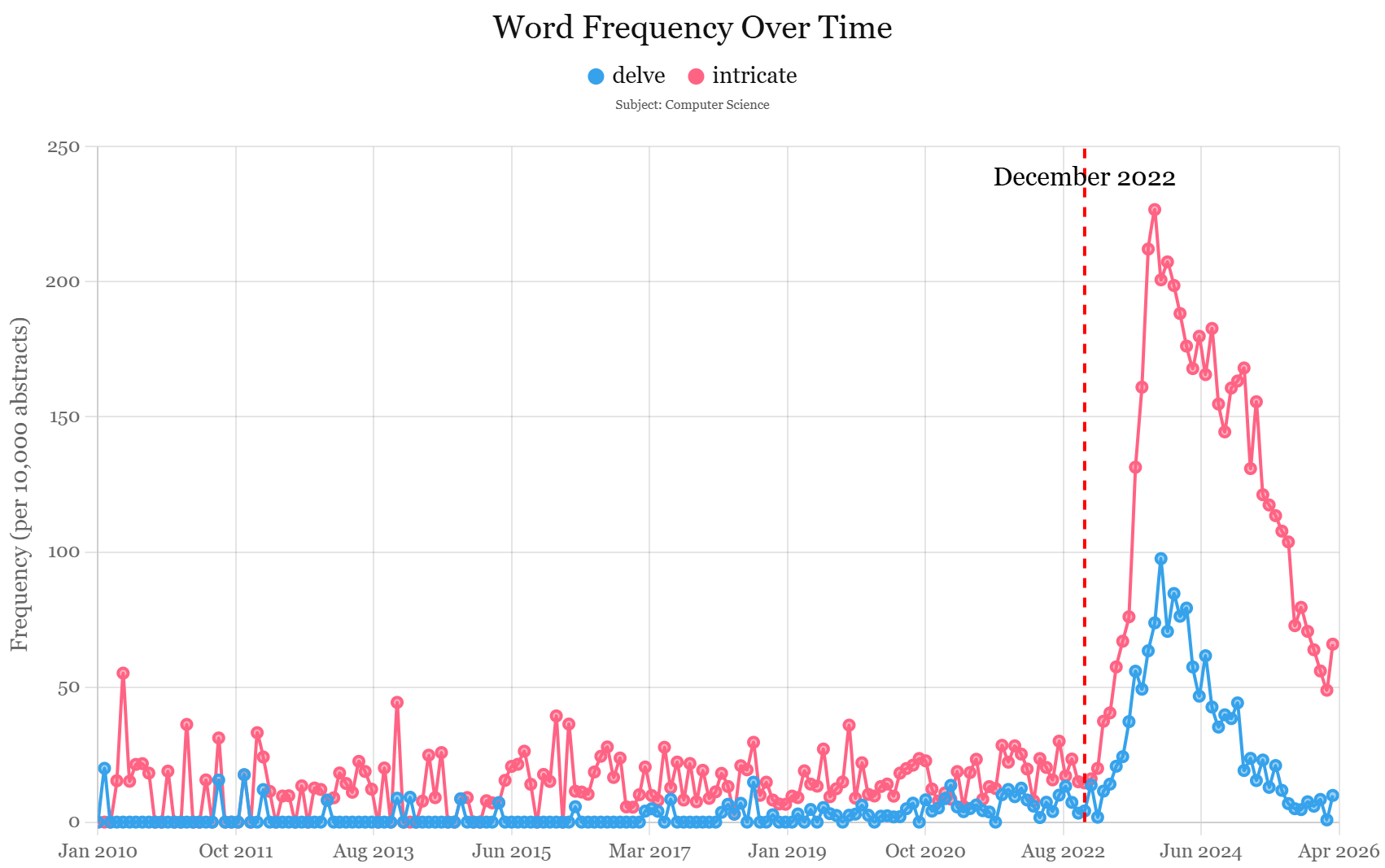}
        \caption{Word frequency in CS abstracts.}
        \label{plot_the_of_cs}
    \end{subfigure}
        \begin{subfigure}{\columnwidth}
        \centering
        \includegraphics[width=0.95\columnwidth]{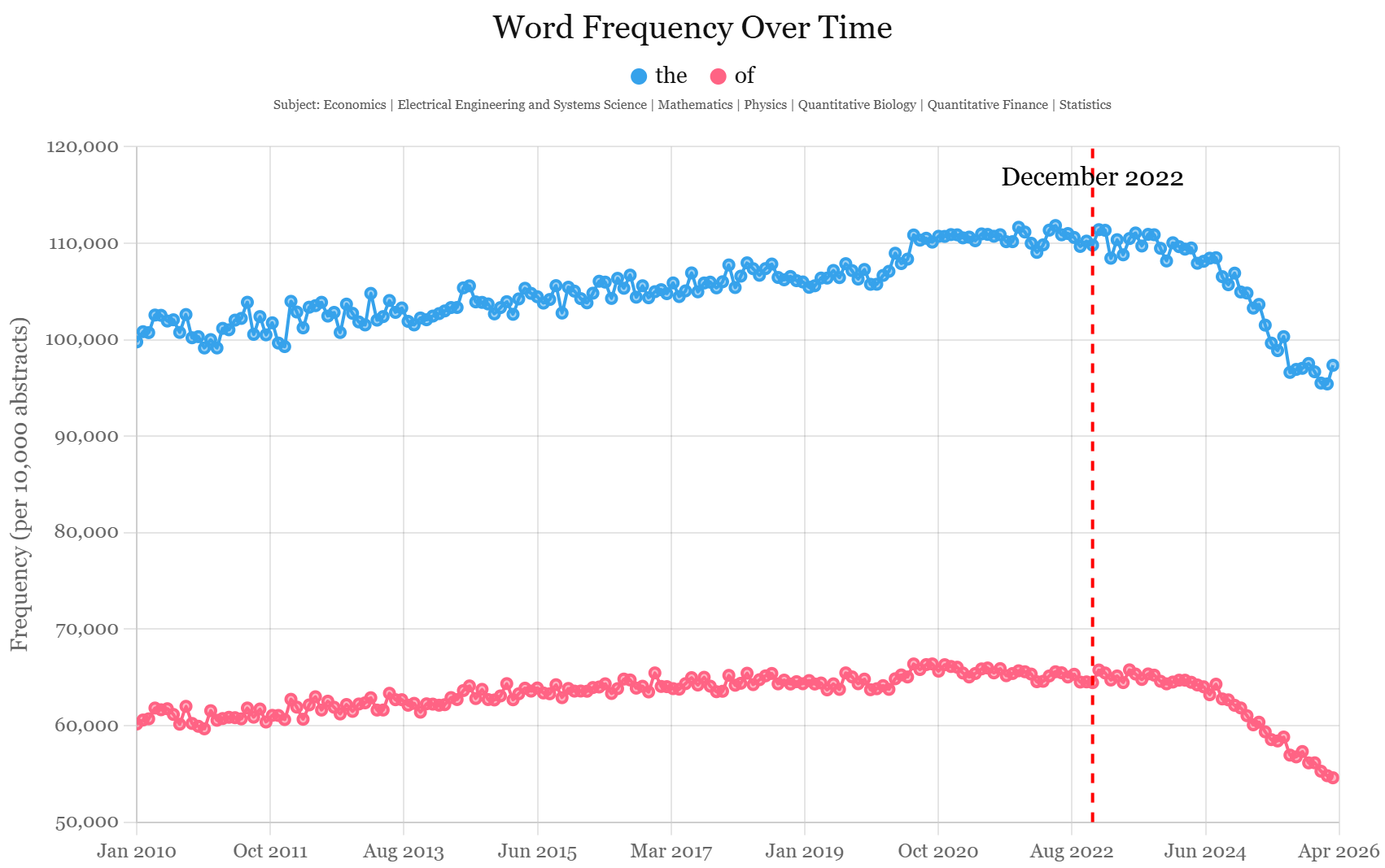}
        \includegraphics[width=0.95\columnwidth]{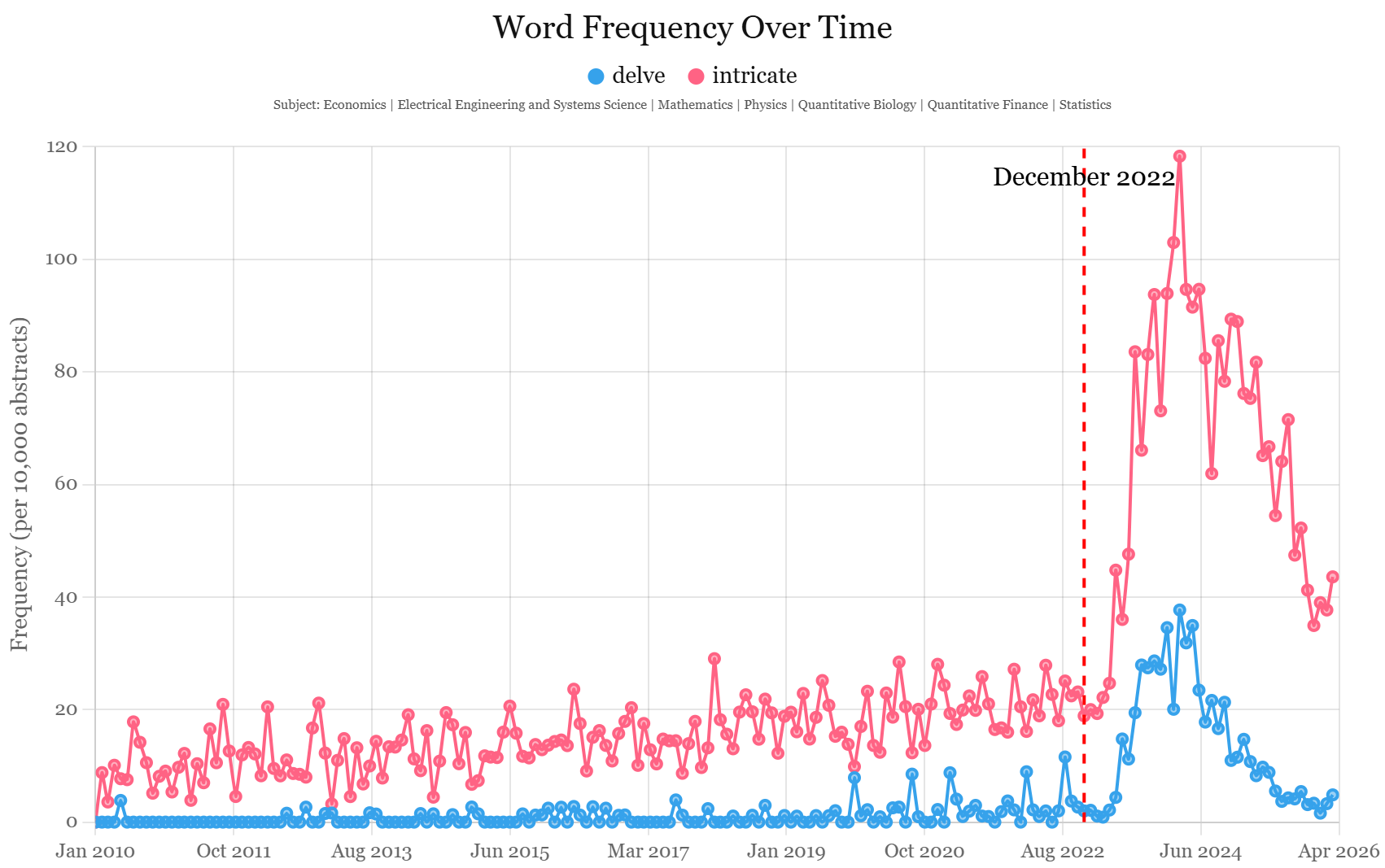}
        \caption{Word frequency in non-CS abstracts.}
        \label{plot_the_of_non_cs}
    \end{subfigure}
    \caption{Supplementary comparison results.}
    \label{the_of_cs_non_cs}
\end{figure}

\clearpage

\begin{table*}[t]
\centering
\caption{Classification results. Prompt: long, Classifier: GPT-2, Model Type: old.}
\label{tab:long_GPT-2_old}
\begin{tabular}{lccccccc}
\toprule
\textbf{GPT} & $\blacksquare$ & $\blacksquare$ & $\blacksquare$ & $\blacksquare$ & \textcolor{gray!30}{$\cdot$} & \textcolor{gray!30}{$\cdot$} & \textcolor{gray!30}{$\cdot$} \\
\textbf{DeepSeek} & $\blacksquare$ & $\blacksquare$ & \textcolor{gray!30}{$\cdot$} & \textcolor{gray!30}{$\cdot$} & $\blacksquare$ & $\blacksquare$ & \textcolor{gray!30}{$\cdot$} \\
\textbf{Gemini} & $\blacksquare$ & \textcolor{gray!30}{$\cdot$} & $\blacksquare$ & \textcolor{gray!30}{$\cdot$} & $\blacksquare$ & \textcolor{gray!30}{$\cdot$} & $\blacksquare$ \\
\textbf{Claude} & $\blacksquare$ & \textcolor{gray!30}{$\cdot$} & \textcolor{gray!30}{$\cdot$} & $\blacksquare$ & \textcolor{gray!30}{$\cdot$} & $\blacksquare$ & $\blacksquare$ \\
\midrule
Accuracy & 78.2\% & 87.1\% & 89.0\% & \textbf{94.8}\% & 71.7\% & 90.1\% & 89.8\% \\
\bottomrule
\end{tabular}
\end{table*}

\begin{table*}[t]
\centering
\caption{Classification results. Prompt: long, Classifier: GPT-2, Model Type: new.}
\label{tab:long_GPT-2_new}
\begin{tabular}{lccccccc}
\toprule
\textbf{GPT} & $\blacksquare$ & $\blacksquare$ & $\blacksquare$ & $\blacksquare$ & \textcolor{gray!30}{$\cdot$} & \textcolor{gray!30}{$\cdot$} & \textcolor{gray!30}{$\cdot$} \\
\textbf{DeepSeek} & $\blacksquare$ & $\blacksquare$ & \textcolor{gray!30}{$\cdot$} & \textcolor{gray!30}{$\cdot$} & $\blacksquare$ & $\blacksquare$ & \textcolor{gray!30}{$\cdot$} \\
\textbf{Gemini} & $\blacksquare$ & \textcolor{gray!30}{$\cdot$} & $\blacksquare$ & \textcolor{gray!30}{$\cdot$} & $\blacksquare$ & \textcolor{gray!30}{$\cdot$} & $\blacksquare$ \\
\textbf{Claude} & $\blacksquare$ & \textcolor{gray!30}{$\cdot$} & \textcolor{gray!30}{$\cdot$} & $\blacksquare$ & \textcolor{gray!30}{$\cdot$} & $\blacksquare$ & $\blacksquare$ \\
\midrule
Accuracy & 65.2\% & 84.6\% & \textbf{90.1}\% & 81.9\% & 77.7\% & 62.9\% & 82.7\% \\
\bottomrule
\end{tabular}
\end{table*}

\begin{table*}[t]
\centering
\caption{Classification results. Prompt: long, Classifier: T5, Model Type: old.}
\label{tab:long_T5_old}
\begin{tabular}{lccccccc}
\toprule
\textbf{GPT} & $\blacksquare$ & $\blacksquare$ & $\blacksquare$ & $\blacksquare$ & \textcolor{gray!30}{$\cdot$} & \textcolor{gray!30}{$\cdot$} & \textcolor{gray!30}{$\cdot$} \\
\textbf{DeepSeek} & $\blacksquare$ & $\blacksquare$ & \textcolor{gray!30}{$\cdot$} & \textcolor{gray!30}{$\cdot$} & $\blacksquare$ & $\blacksquare$ & \textcolor{gray!30}{$\cdot$} \\
\textbf{Gemini} & $\blacksquare$ & \textcolor{gray!30}{$\cdot$} & $\blacksquare$ & \textcolor{gray!30}{$\cdot$} & $\blacksquare$ & \textcolor{gray!30}{$\cdot$} & $\blacksquare$ \\
\textbf{Claude} & $\blacksquare$ & \textcolor{gray!30}{$\cdot$} & \textcolor{gray!30}{$\cdot$} & $\blacksquare$ & \textcolor{gray!30}{$\cdot$} & $\blacksquare$ & $\blacksquare$ \\
\midrule
Accuracy & 72.8\% & 90.7\% & 91.2\% & \textbf{94.8}\% & 73.4\% & 91.2\% & 91.8\% \\
\bottomrule
\end{tabular}
\end{table*}

\begin{table*}[t]
\centering
\caption{Classification results. Prompt: long, Classifier: T5, Model Type: new.}
\label{tab:long_T5_new}
\begin{tabular}{lccccccc}
\toprule
\textbf{GPT} & $\blacksquare$ & $\blacksquare$ & $\blacksquare$ & $\blacksquare$ & \textcolor{gray!30}{$\cdot$} & \textcolor{gray!30}{$\cdot$} & \textcolor{gray!30}{$\cdot$} \\
\textbf{DeepSeek} & $\blacksquare$ & $\blacksquare$ & \textcolor{gray!30}{$\cdot$} & \textcolor{gray!30}{$\cdot$} & $\blacksquare$ & $\blacksquare$ & \textcolor{gray!30}{$\cdot$} \\
\textbf{Gemini} & $\blacksquare$ & \textcolor{gray!30}{$\cdot$} & $\blacksquare$ & \textcolor{gray!30}{$\cdot$} & $\blacksquare$ & \textcolor{gray!30}{$\cdot$} & $\blacksquare$ \\
\textbf{Claude} & $\blacksquare$ & \textcolor{gray!30}{$\cdot$} & \textcolor{gray!30}{$\cdot$} & $\blacksquare$ & \textcolor{gray!30}{$\cdot$} & $\blacksquare$ & $\blacksquare$ \\
\midrule
Accuracy & 57.3\% & 88.5\% & \textbf{93.1}\% & 84.1\% & 83.8\% & 58.5\% & 82.4\% \\
\bottomrule
\end{tabular}
\end{table*}

\begin{table*}[t]
\centering
\caption{Classification results. Prompt: short, Classifier: BERT, Model Type: old.}
\label{tab:short_BERT_old}
\begin{tabular}{lccccccc}
\toprule
\textbf{GPT} & $\blacksquare$ & $\blacksquare$ & $\blacksquare$ & $\blacksquare$ & \textcolor{gray!30}{$\cdot$} & \textcolor{gray!30}{$\cdot$} & \textcolor{gray!30}{$\cdot$} \\
\textbf{DeepSeek} & $\blacksquare$ & $\blacksquare$ & \textcolor{gray!30}{$\cdot$} & \textcolor{gray!30}{$\cdot$} & $\blacksquare$ & $\blacksquare$ & \textcolor{gray!30}{$\cdot$} \\
\textbf{Gemini} & $\blacksquare$ & \textcolor{gray!30}{$\cdot$} & $\blacksquare$ & \textcolor{gray!30}{$\cdot$} & $\blacksquare$ & \textcolor{gray!30}{$\cdot$} & $\blacksquare$ \\
\textbf{Claude} & $\blacksquare$ & \textcolor{gray!30}{$\cdot$} & \textcolor{gray!30}{$\cdot$} & $\blacksquare$ & \textcolor{gray!30}{$\cdot$} & $\blacksquare$ & $\blacksquare$ \\
\midrule
Accuracy & 73.1\% & 82.1\% & 89.3\% & 91.5\% & 71.2\% & 89.3\% & \textbf{93.4}\% \\
\bottomrule
\end{tabular}
\end{table*}

\begin{table*}[t]
\centering
\caption{Classification results. Prompt: short, Classifier: BERT, Model Type: new.}
\label{tab:short_BERT_new}
\begin{tabular}{lccccccc}
\toprule
\textbf{GPT} & $\blacksquare$ & $\blacksquare$ & $\blacksquare$ & $\blacksquare$ & \textcolor{gray!30}{$\cdot$} & \textcolor{gray!30}{$\cdot$} & \textcolor{gray!30}{$\cdot$} \\
\textbf{DeepSeek} & $\blacksquare$ & $\blacksquare$ & \textcolor{gray!30}{$\cdot$} & \textcolor{gray!30}{$\cdot$} & $\blacksquare$ & $\blacksquare$ & \textcolor{gray!30}{$\cdot$} \\
\textbf{Gemini} & $\blacksquare$ & \textcolor{gray!30}{$\cdot$} & $\blacksquare$ & \textcolor{gray!30}{$\cdot$} & $\blacksquare$ & \textcolor{gray!30}{$\cdot$} & $\blacksquare$ \\
\textbf{Claude} & $\blacksquare$ & \textcolor{gray!30}{$\cdot$} & \textcolor{gray!30}{$\cdot$} & $\blacksquare$ & \textcolor{gray!30}{$\cdot$} & $\blacksquare$ & $\blacksquare$ \\
\midrule
Accuracy & 65.5\% & 80.8\% & \textbf{89.8}\% & 83.8\% & 87.6\% & 76.6\% & 86.8\% \\
\bottomrule
\end{tabular}
\end{table*}

\begin{table*}[t]
\centering
\caption{Classification results. Prompt: short, Classifier: GPT-2, Model Type: old.}
\label{tab:short_GPT-2_old}
\begin{tabular}{lccccccc}
\toprule
\textbf{GPT} & $\blacksquare$ & $\blacksquare$ & $\blacksquare$ & $\blacksquare$ & \textcolor{gray!30}{$\cdot$} & \textcolor{gray!30}{$\cdot$} & \textcolor{gray!30}{$\cdot$} \\
\textbf{DeepSeek} & $\blacksquare$ & $\blacksquare$ & \textcolor{gray!30}{$\cdot$} & \textcolor{gray!30}{$\cdot$} & $\blacksquare$ & $\blacksquare$ & \textcolor{gray!30}{$\cdot$} \\
\textbf{Gemini} & $\blacksquare$ & \textcolor{gray!30}{$\cdot$} & $\blacksquare$ & \textcolor{gray!30}{$\cdot$} & $\blacksquare$ & \textcolor{gray!30}{$\cdot$} & $\blacksquare$ \\
\textbf{Claude} & $\blacksquare$ & \textcolor{gray!30}{$\cdot$} & \textcolor{gray!30}{$\cdot$} & $\blacksquare$ & \textcolor{gray!30}{$\cdot$} & $\blacksquare$ & $\blacksquare$ \\
\midrule
Accuracy & 69.0\% & 80.2\% & 88.2\% & 89.6\% & 73.1\% & 85.7\% & \textbf{90.9}\% \\
\bottomrule
\end{tabular}
\end{table*}

\begin{table*}[t]
\centering
\caption{Classification results. Prompt: short, Classifier: GPT-2, Model Type: new.}
\label{tab:short_GPT-2_new}
\begin{tabular}{lccccccc}
\toprule
\textbf{GPT} & $\blacksquare$ & $\blacksquare$ & $\blacksquare$ & $\blacksquare$ & \textcolor{gray!30}{$\cdot$} & \textcolor{gray!30}{$\cdot$} & \textcolor{gray!30}{$\cdot$} \\
\textbf{DeepSeek} & $\blacksquare$ & $\blacksquare$ & \textcolor{gray!30}{$\cdot$} & \textcolor{gray!30}{$\cdot$} & $\blacksquare$ & $\blacksquare$ & \textcolor{gray!30}{$\cdot$} \\
\textbf{Gemini} & $\blacksquare$ & \textcolor{gray!30}{$\cdot$} & $\blacksquare$ & \textcolor{gray!30}{$\cdot$} & $\blacksquare$ & \textcolor{gray!30}{$\cdot$} & $\blacksquare$ \\
\textbf{Claude} & $\blacksquare$ & \textcolor{gray!30}{$\cdot$} & \textcolor{gray!30}{$\cdot$} & $\blacksquare$ & \textcolor{gray!30}{$\cdot$} & $\blacksquare$ & $\blacksquare$ \\
\midrule
Accuracy & 58.2\% & 76.6\% & \textbf{89.6}\% & 83.0\% & 80.8\% & 70.6\% & 83.0\% \\
\bottomrule
\end{tabular}
\end{table*}

\begin{table*}[t]
\centering
\caption{Classification results. Prompt: short, Classifier: T5, Model Type: old.}
\label{tab:short_T5_old}
\begin{tabular}{lccccccc}
\toprule
\textbf{GPT} & $\blacksquare$ & $\blacksquare$ & $\blacksquare$ & $\blacksquare$ & \textcolor{gray!30}{$\cdot$} & \textcolor{gray!30}{$\cdot$} & \textcolor{gray!30}{$\cdot$} \\
\textbf{DeepSeek} & $\blacksquare$ & $\blacksquare$ & \textcolor{gray!30}{$\cdot$} & \textcolor{gray!30}{$\cdot$} & $\blacksquare$ & $\blacksquare$ & \textcolor{gray!30}{$\cdot$} \\
\textbf{Gemini} & $\blacksquare$ & \textcolor{gray!30}{$\cdot$} & $\blacksquare$ & \textcolor{gray!30}{$\cdot$} & $\blacksquare$ & \textcolor{gray!30}{$\cdot$} & $\blacksquare$ \\
\textbf{Claude} & $\blacksquare$ & \textcolor{gray!30}{$\cdot$} & \textcolor{gray!30}{$\cdot$} & $\blacksquare$ & \textcolor{gray!30}{$\cdot$} & $\blacksquare$ & $\blacksquare$ \\
\midrule
Accuracy & 70.9\% & 83.5\% & 91.2\% & \textbf{92.9}\% & 69.8\% & 86.3\% & \textbf{92.9}\% \\
\bottomrule
\end{tabular}
\end{table*}

\begin{table*}[t]
\centering
\caption{Classification results. Prompt: short, Classifier: T5, Model Type: new.}
\label{tab:short_T5_new}
\begin{tabular}{lccccccc}
\toprule
\textbf{GPT} & $\blacksquare$ & $\blacksquare$ & $\blacksquare$ & $\blacksquare$ & \textcolor{gray!30}{$\cdot$} & \textcolor{gray!30}{$\cdot$} & \textcolor{gray!30}{$\cdot$} \\
\textbf{DeepSeek} & $\blacksquare$ & $\blacksquare$ & \textcolor{gray!30}{$\cdot$} & \textcolor{gray!30}{$\cdot$} & $\blacksquare$ & $\blacksquare$ & \textcolor{gray!30}{$\cdot$} \\
\textbf{Gemini} & $\blacksquare$ & \textcolor{gray!30}{$\cdot$} & $\blacksquare$ & \textcolor{gray!30}{$\cdot$} & $\blacksquare$ & \textcolor{gray!30}{$\cdot$} & $\blacksquare$ \\
\textbf{Claude} & $\blacksquare$ & \textcolor{gray!30}{$\cdot$} & \textcolor{gray!30}{$\cdot$} & $\blacksquare$ & \textcolor{gray!30}{$\cdot$} & $\blacksquare$ & $\blacksquare$ \\
\midrule
Accuracy & 63.5\% & 82.7\% & \textbf{90.4}\% & 85.4\% & 83.2\% & 64.6\% & 84.3\% \\
\bottomrule
\end{tabular}
\end{table*}

\begin{table*}[t]
\centering
\caption{Summary of Binary Classification Accuracy (\%) across different Model Types, Prompts, and Classifiers. \\ 
\small{\textbf{Abbreviations:} G: GPT, D: DeepSeek, Ge: Gemini, C: Claude. Columns represent the ensemble combination.}}
\label{tab:summary_classification}
\resizebox{\textwidth}{!}{%
\begin{tabular}{lllccccccc}
\toprule
\multirow{2}{*}{\textbf{Model Type}} & \multirow{2}{*}{\textbf{Prompt}} & \multirow{2}{*}{\textbf{Classifier}} & \textbf{All} & \textbf{G + D} & \textbf{G + Ge} & \textbf{G + C} & \textbf{D + Ge} & \textbf{D + C} & \textbf{Ge + C} \\
 &  &  & (G+D+Ge+C) &  &  &  &  &  &  \\
\midrule
\multirow{6}{*}{\textbf{New}} & \multirow{3}{*}{Long} & BERT & 63.0 & 83.0 & \textbf{90.4} & 84.9 & 83.8 & 70.9 & 88.2 \\
 &  & GPT-2 & 65.2 & 84.6 & \textbf{90.1} & 81.9 & 77.7 & 62.9 & 82.7 \\
 &  & T5 & 57.3 & 88.5 & \textbf{93.1} & 84.1 & 83.8 & 58.5 & 82.4 \\
\cmidrule{2-10}
 & \multirow{3}{*}{Short} & BERT & 65.5 & 80.8 & \textbf{89.8} & 83.8 & 87.6 & 76.6 & 86.8 \\
 &  & GPT-2 & 58.2 & 76.6 & \textbf{89.6} & 83.0 & 80.8 & 70.6 & 83.0 \\
 &  & T5 & 63.5 & 82.7 & \textbf{90.4} & 85.4 & 83.2 & 64.6 & 84.3 \\
\midrule
\multirow{6}{*}{\textbf{Old}} & \multirow{3}{*}{Long} & BERT & 78.6 & 90.1 & 90.1 & \textbf{95.3} & 76.9 & 93.4 & 91.8 \\
 &  & GPT-2 & 78.2 & 87.1 & 89.0 & \textbf{94.8} & 71.7 & 90.1 & 89.8 \\
 &  & T5 & 72.8 & 90.7 & 91.2 & \textbf{94.8} & 73.4 & 91.2 & 91.8 \\
\cmidrule{2-10}
 & \multirow{3}{*}{Short} & BERT & 73.1 & 82.1 & 89.3 & 91.5 & 71.2 & 89.3 & \textbf{93.4} \\
 &  & GPT-2 & 69.0 & 80.2 & 88.2 & 89.6 & 73.1 & 85.7 & \textbf{90.9} \\
 &  & T5 & 70.9 & 83.5 & 91.2 & \textbf{92.9} & 69.8 & 86.3 & \textbf{92.9} \\
\bottomrule
\end{tabular}%
}
\end{table*}

\newpage

\begin{figure*}[t]
    \begin{subfigure}{\textwidth}
        \centering
        \includegraphics[width=\textwidth]{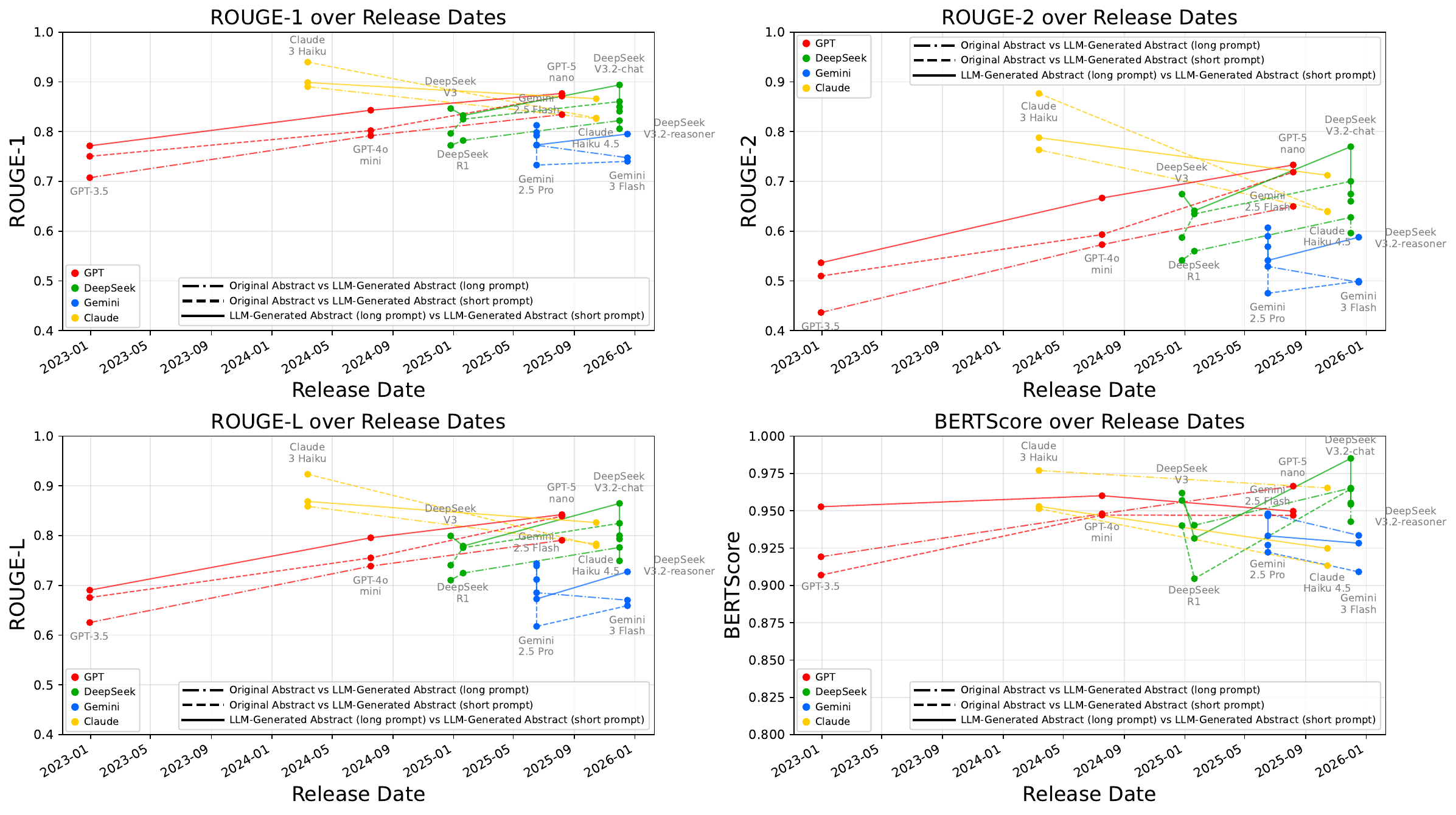}
        \caption{Similarity between the processed texts of the two prompts and their similarity to the real abstracts.}
        \label{abstract_metrics_grid_v2}
    \end{subfigure}
        \begin{subfigure}{\textwidth}
        \centering
        \includegraphics[width=\textwidth]{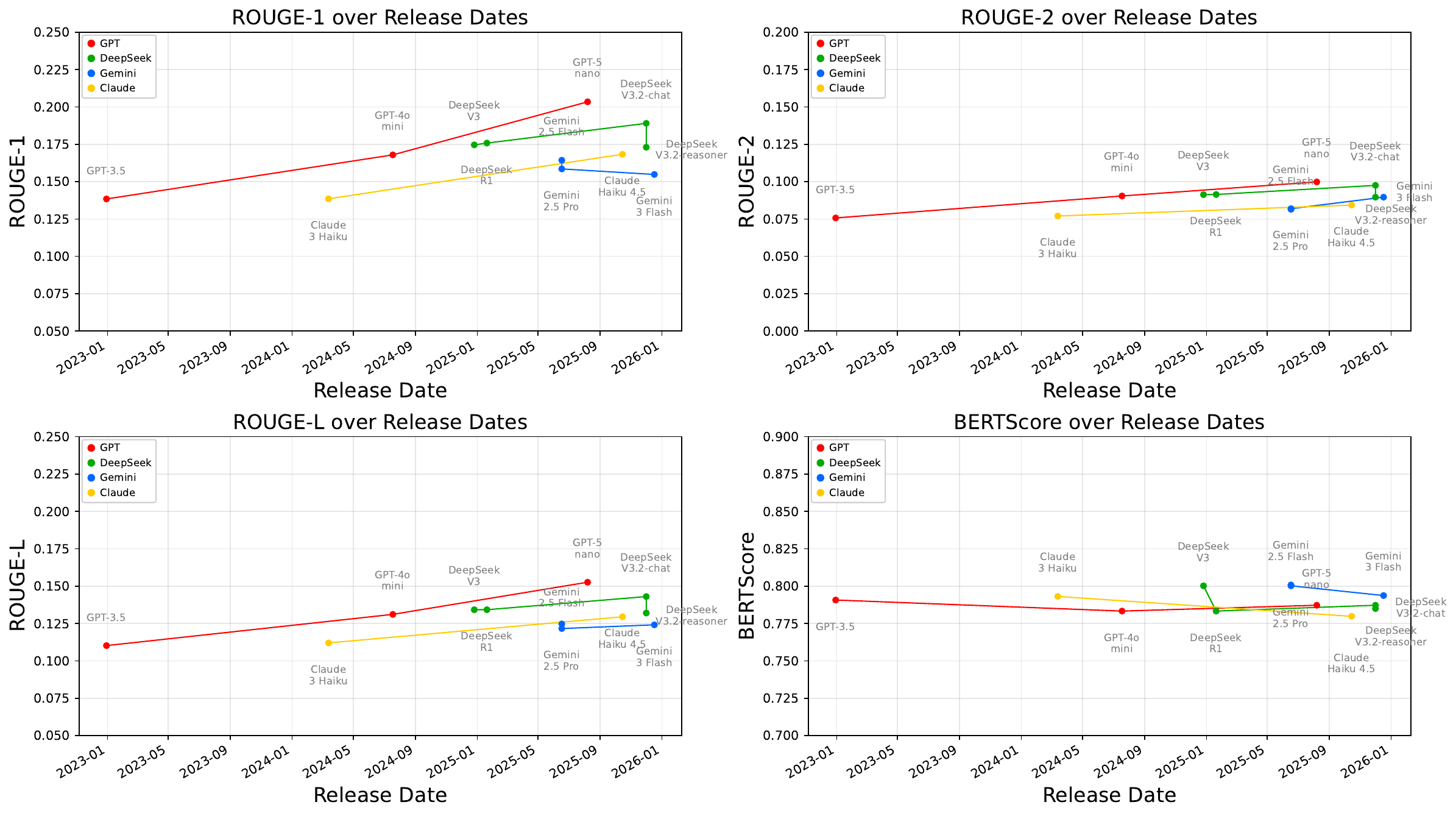}
        \caption{Similarity between the titles generated by LLMs and the real titles.}
        \label{title_metrics_grid_v2}
    \end{subfigure}
    \caption{Comparison Results of Text Similarity. The x-axis represents the model release dates.}
    \label{metrics_grid_v2}
\end{figure*}

\begin{table*}[t]
  \centering
  \caption{Comparison of 7-class classification performance across BERT, GPT-2, T5, and LLM2Vec models. Metrics are reported in percentage (\%). P denotes Precision, R denotes Recall.}
  \label{tab:combined_classification}
  \resizebox{\textwidth}{!}{%
  \begin{tabular}{lcccccccccccc}
    \toprule
    \multirow{2}{*}{\textbf{Class}} & \multicolumn{3}{c}{\textbf{BERT}} & \multicolumn{3}{c}{\textbf{GPT-2}} & \multicolumn{3}{c}{\textbf{T5}} & \multicolumn{3}{c}{\textbf{LLM2Vec}} \\
    \cmidrule(lr){2-4} \cmidrule(lr){5-7} \cmidrule(lr){8-10} \cmidrule(lr){11-13}
     & \textbf{P} & \textbf{R} & \textbf{F1} & \textbf{P} & \textbf{R} & \textbf{F1} & \textbf{P} & \textbf{R} & \textbf{F1} & \textbf{P} & \textbf{R} & \textbf{F1} \\
    \midrule
    Real abstract & \textbf{82.24} & 68.68 & 74.85 & \textbf{74.21} & \textbf{77.47} & \textbf{75.81} & \textbf{85.06} & 71.98 & \textbf{77.98} & \textbf{86.93} & 84.07 & 85.47 \\
    GPT 3.5 & 74.07 & \textbf{76.92} & \textbf{75.47} & 66.50 & 71.98 & 69.13 & 79.62 & 68.68 & 73.75 & 86.41 & \textbf{87.36} & \textbf{86.89} \\
    GPT 4o mini & 64.71 & 72.53 & 68.39 & 58.29 & 67.58 & 62.60 & 60.68 & \textbf{78.02} & 68.27 & 83.05 & 80.77 & 81.89 \\
    GPT 5 nano & 68.45 & 63.19 & 65.71 & 69.14 & 66.48 & 67.79 & 67.50 & 74.18 & 70.68 & 84.24 & 76.37 & 80.12 \\
    DeepSeek & 36.18 & 30.22 & 32.93 & 31.03 & 19.78 & 24.16 & 38.71 & 26.37 & 31.37 & 49.45 & 49.45 & 49.45 \\
    Gemini & 44.56 & 71.98 & 55.04 & 45.57 & 59.34 & 51.55 & 48.55 & 73.63 & 58.52 & 70.21 & 72.53 & 71.35 \\
    Claude & 41.74 & 26.37 & 32.32 & 39.86 & 32.42 & 35.76 & 48.06 & 34.07 & 39.87 & 51.98 & 57.69 & 54.69 \\
    \midrule
    \textbf{Overall Accuracy} & \multicolumn{3}{c}{\textbf{58.56}} & \multicolumn{3}{c}{\textbf{56.44}} & \multicolumn{3}{c}{\textbf{60.99}} & \multicolumn{3}{c}{\textbf{72.61}} \\
    \bottomrule
  \end{tabular}%
  }
\end{table*}

\begin{figure*}[t]
    \centering
    
    \includegraphics[width=0.7\textwidth]{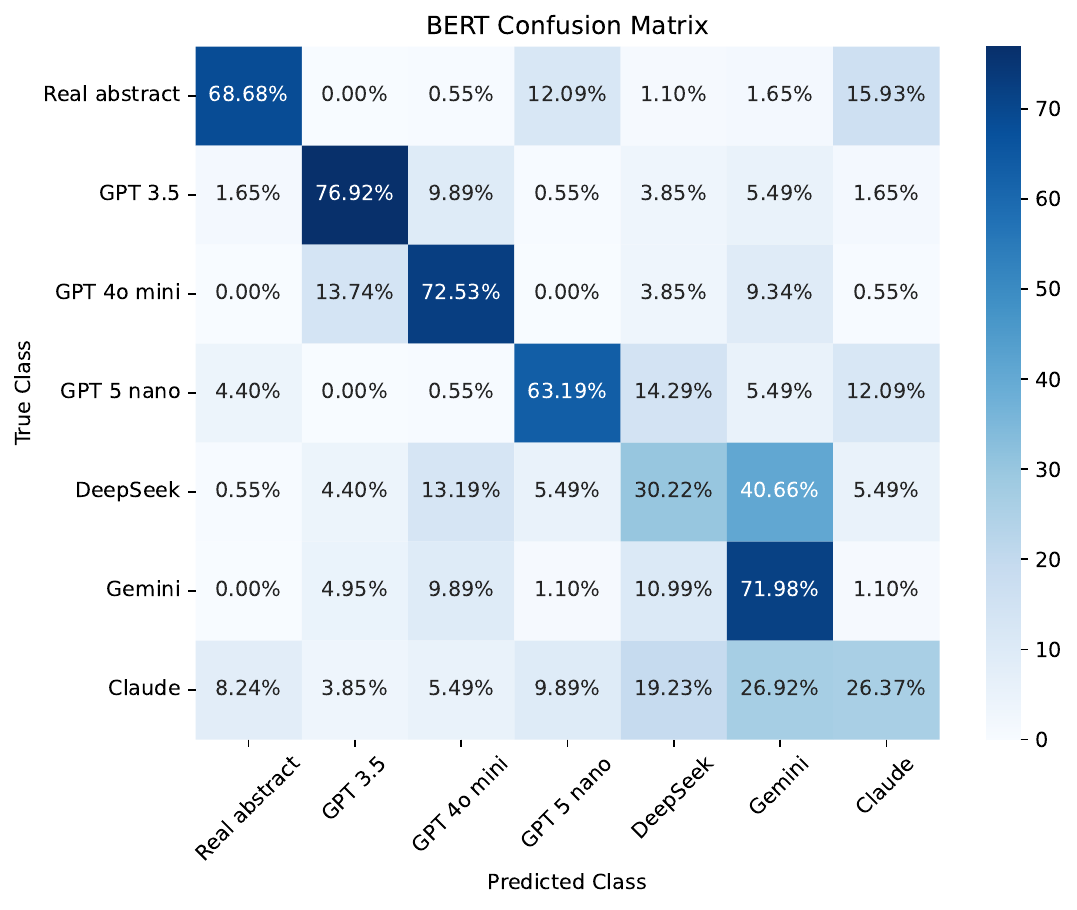}
    \vspace{0.2cm}
    
    \includegraphics[width=0.7\textwidth]{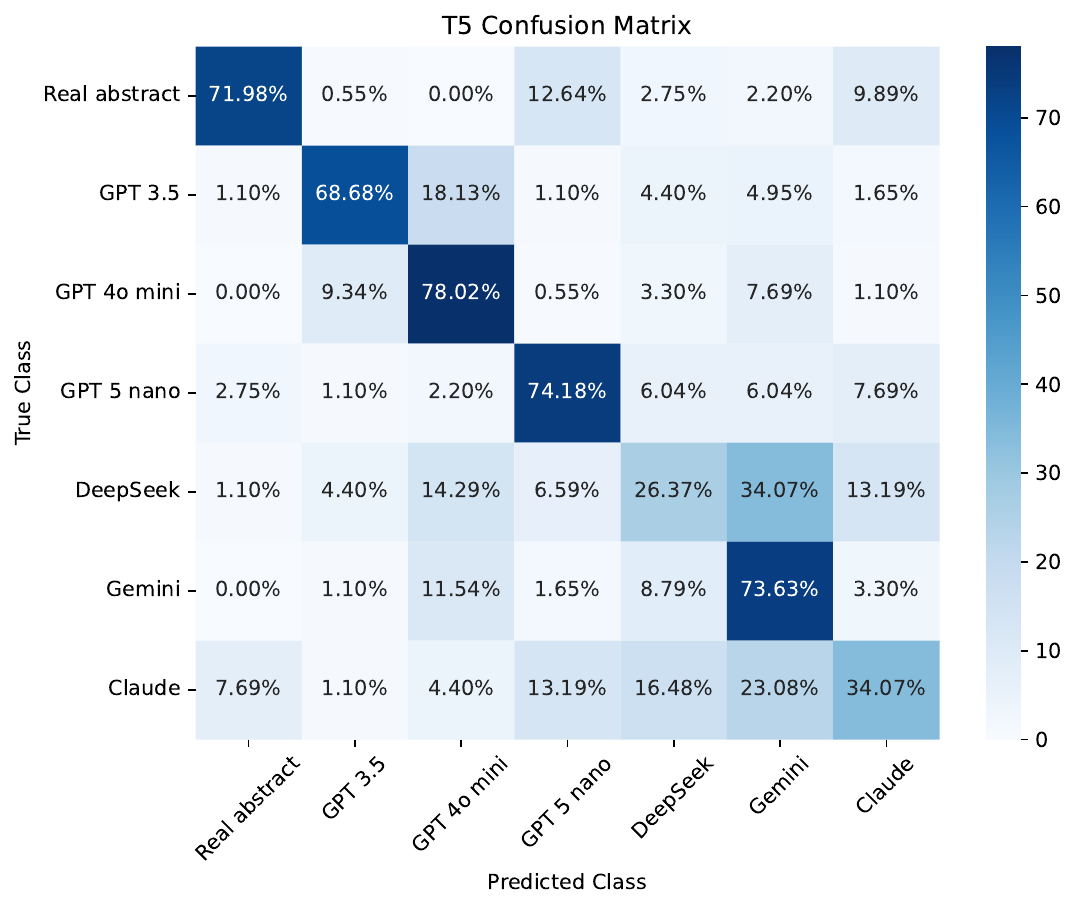}
    
    \caption{Confusion matrix of classification results from the GPT-2-based detector and T5-based detector.}
    \label{confusion_matrix_gpt_t5}
\end{figure*}

\begin{figure*}[t]
    \centering
    
    \includegraphics[width=0.7\textwidth]{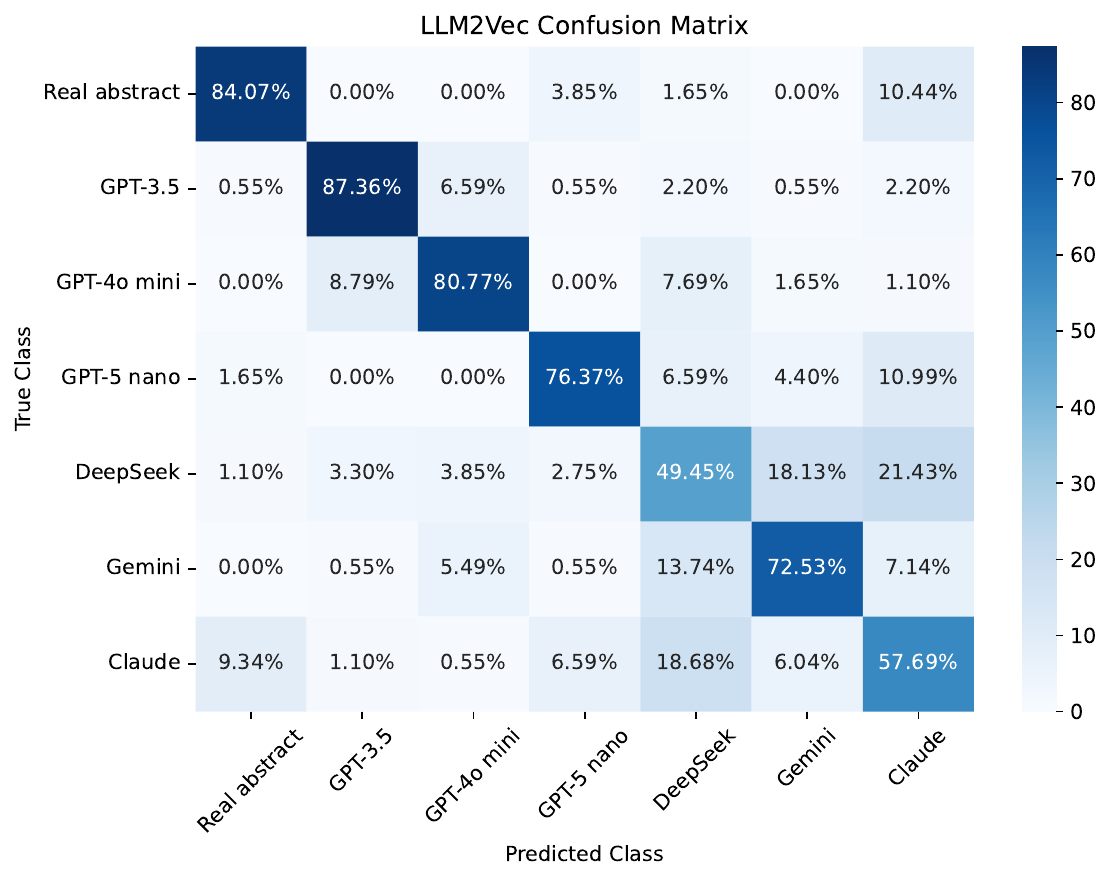}
    
    \caption{Confusion matrix of classification results from the LLM2Vec-based detector.}
    \label{confusion_matrix_llm2vec}
\end{figure*}

\begin{figure*}[t]
  \centering
  \includegraphics[width=\textwidth]{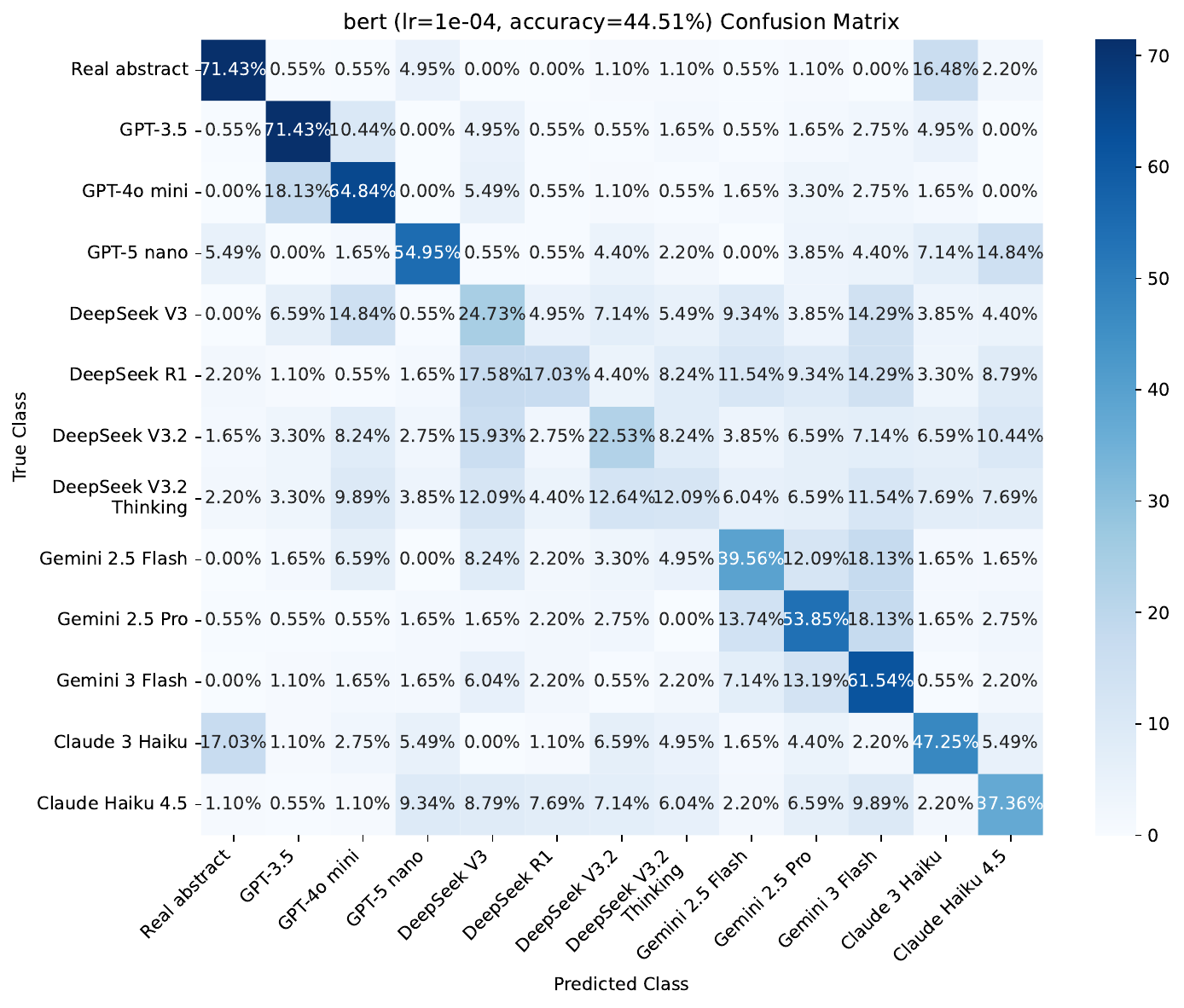}
  \caption{Confusion matrix of BERT (lr=1e-04)}
  \label{fig:confmat-bert}
\end{figure*}

\begin{figure*}[t]
  \centering
  \includegraphics[width=\textwidth]{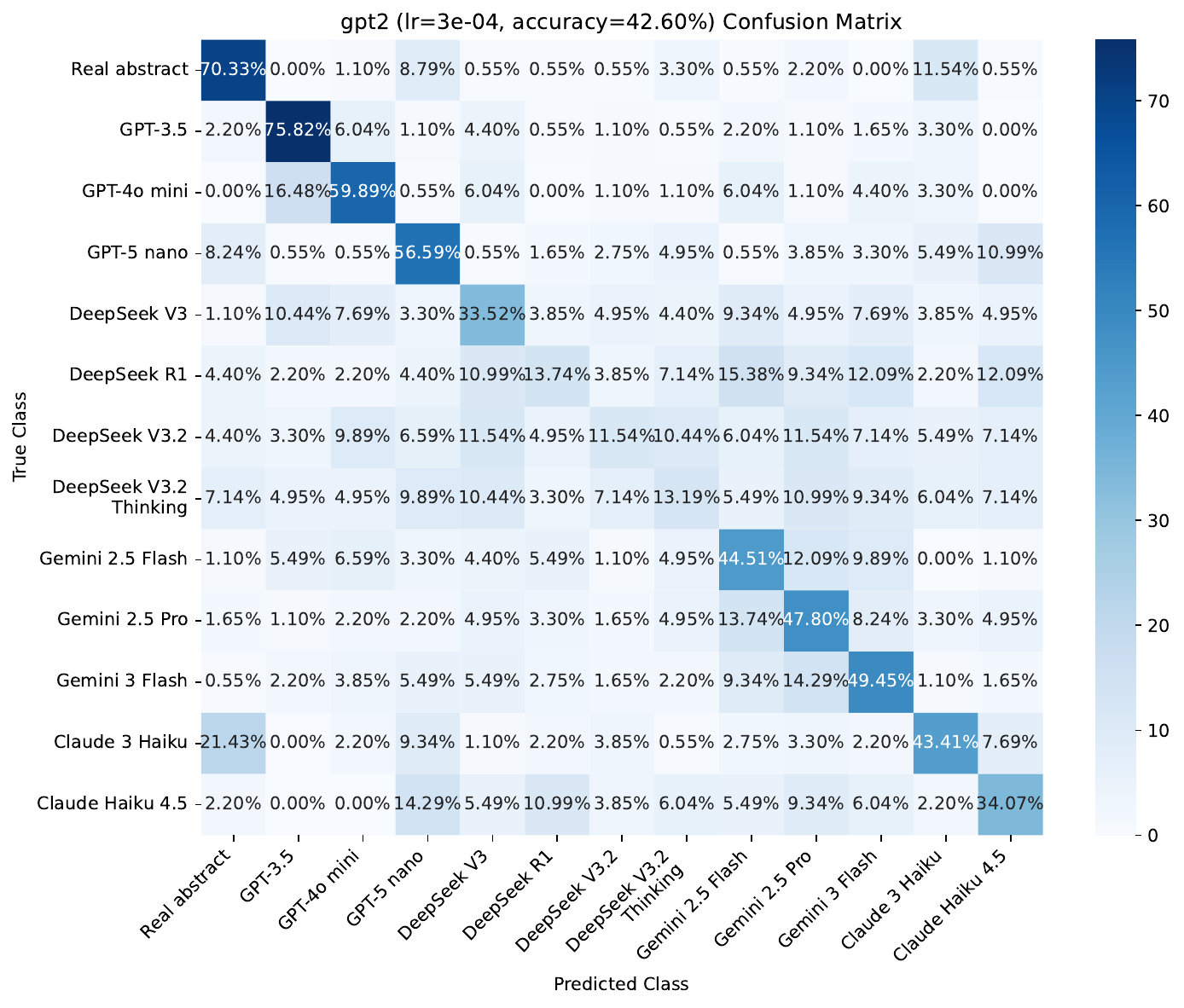}
  \caption{Confusion matrix of GPT-2 (lr=3e-04)}
  \label{fig:confmat-gpt2}
\end{figure*}

\begin{figure*}[t]
  \centering
  \includegraphics[width=\textwidth]{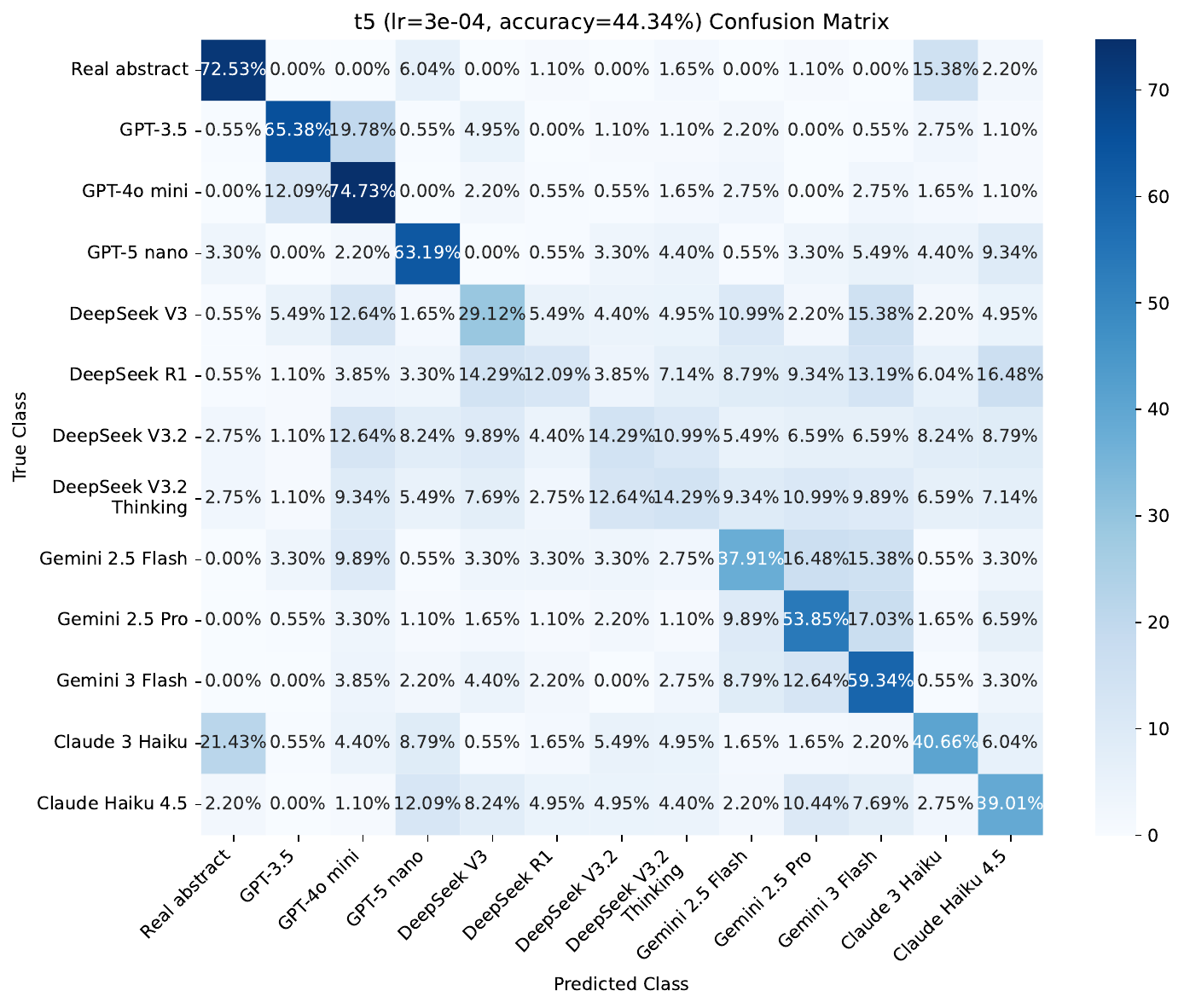}
  \caption{Confusion matrix of T5 (lr=3e-04)}
  \label{fig:confmat-t5}
\end{figure*}

\begin{figure*}[t]
  \centering
  \includegraphics[width=\textwidth]{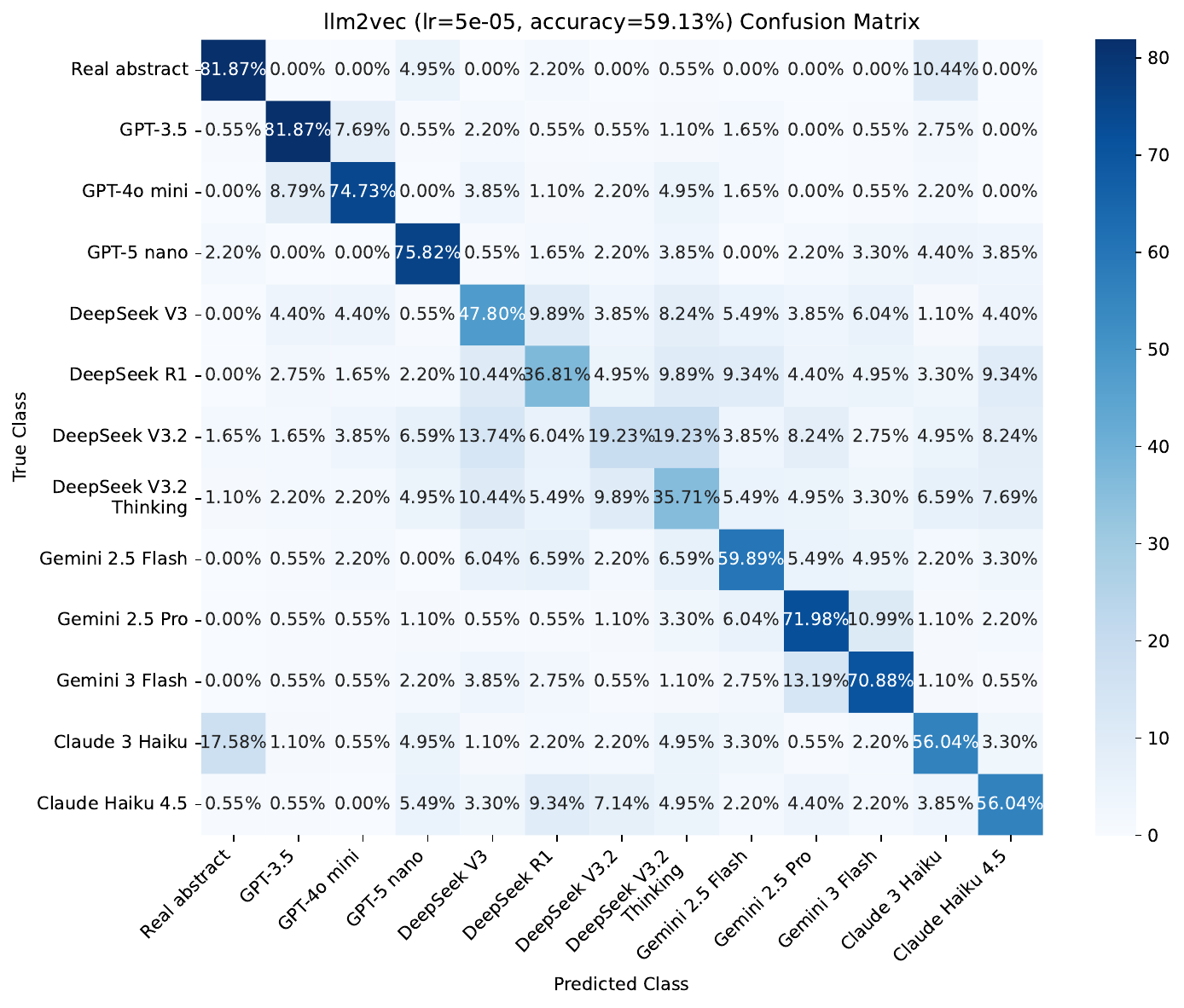}
  \caption{Confusion matrix of LLM2Vec (lr=5e-05)}
  \label{fig:confmat-llm2vec}
\end{figure*}

\begin{figure*}[t]
    \begin{subfigure}{0.98\columnwidth}
        \centering
        \includegraphics[width=\textwidth]{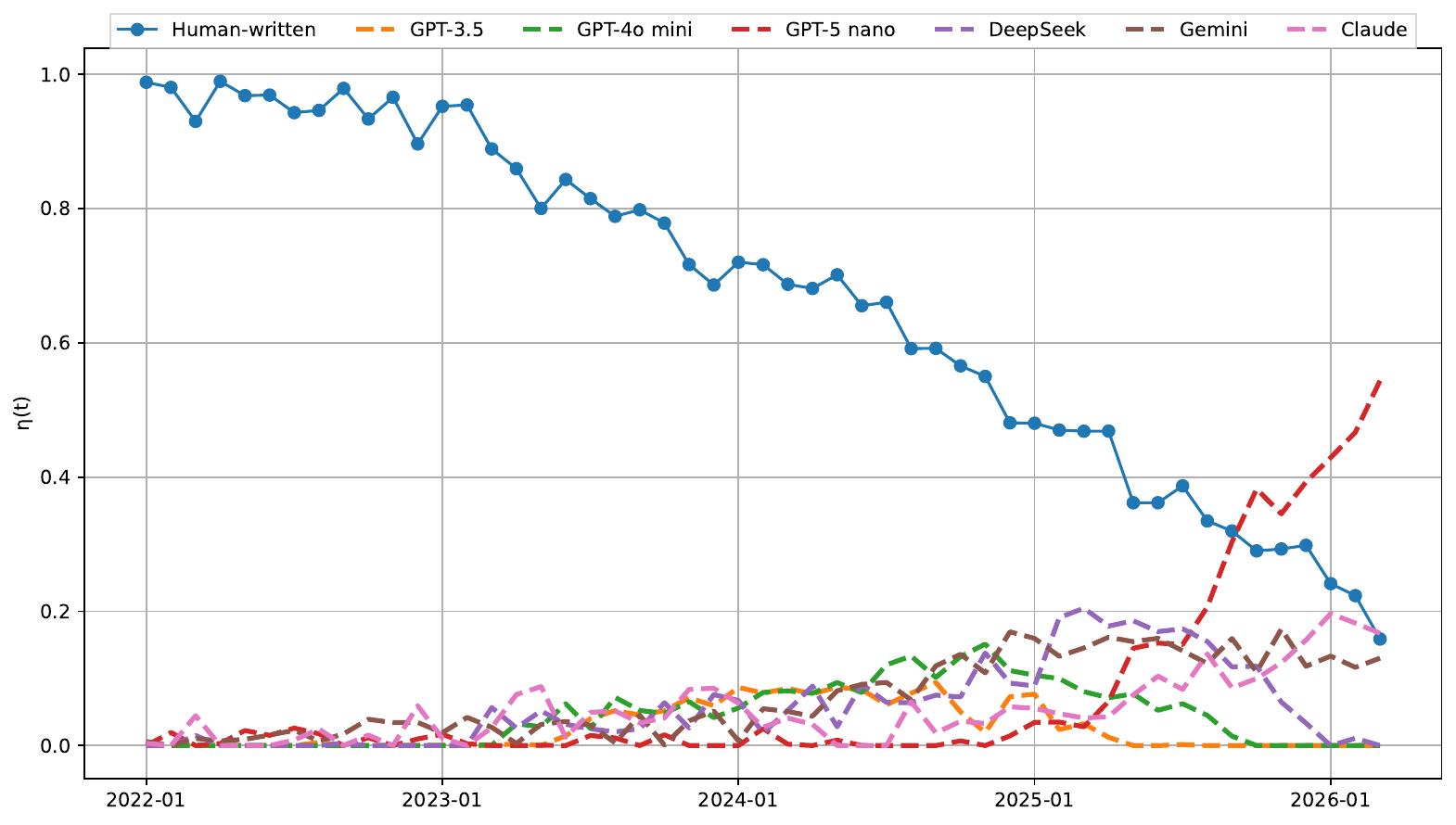}
        \caption{The 695 words used for estimation are drawn from the 1,000 most frequent words in the Google Books Ngram dataset that occur at least 10 times in the 2,000 preprocessed abstracts used for simulation.}
        \label{eta_estimation_10_0_1000}
    \end{subfigure}
    \hfill
    \begin{subfigure}{0.98\columnwidth}
        \centering
        \includegraphics[width=\textwidth]{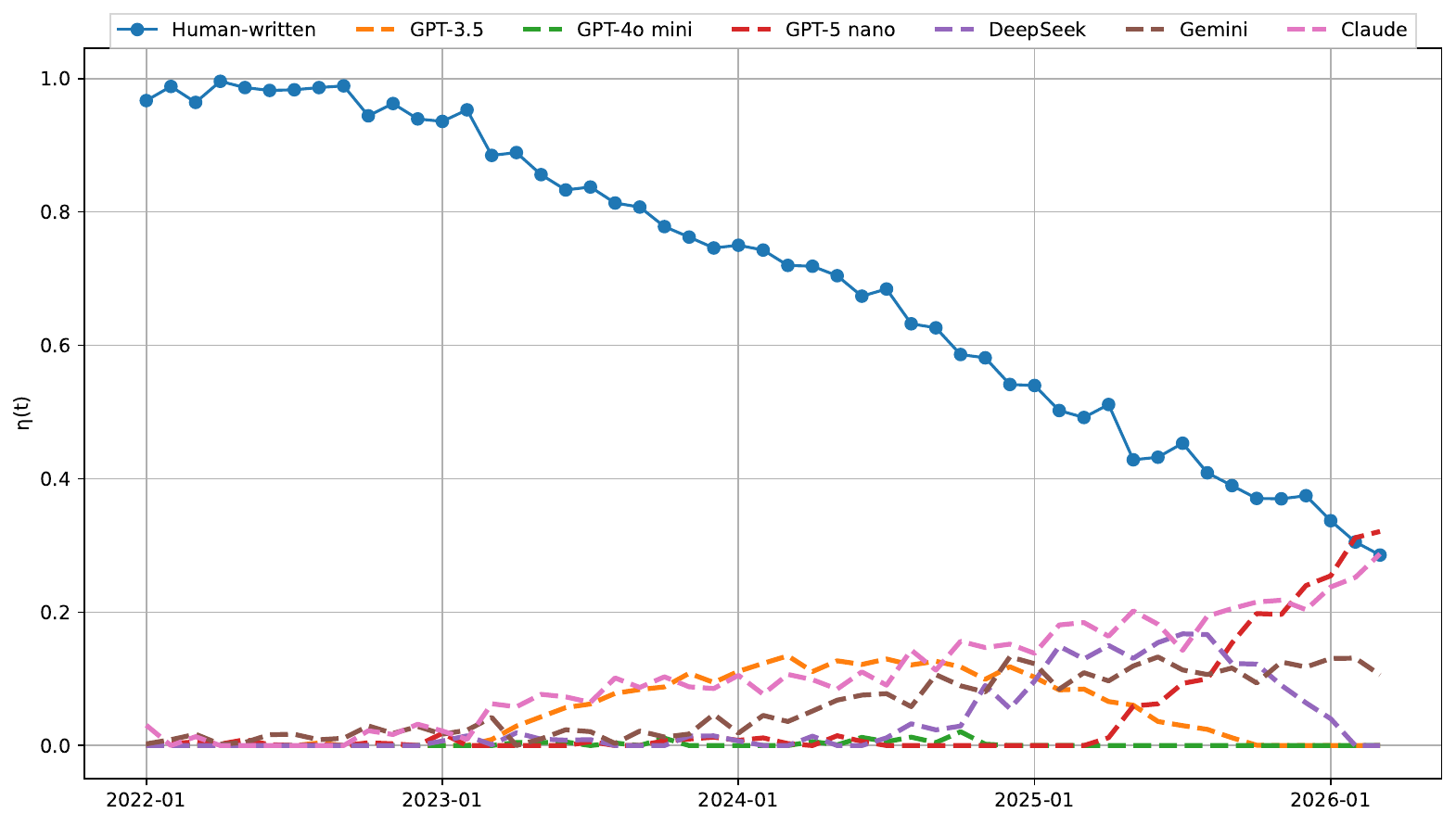}
        \caption{The 2,088 words used for estimation are drawn from the 5,000 most frequent words in the Google Books Ngram dataset that occur at least 10 times in the 2,000 preprocessed abstracts used for simulation.}
        \label{eta_estimation_10_0_5000}
    \end{subfigure}
    \hfill
    \begin{subfigure}{0.98\columnwidth}
        \centering
        \includegraphics[width=\textwidth]{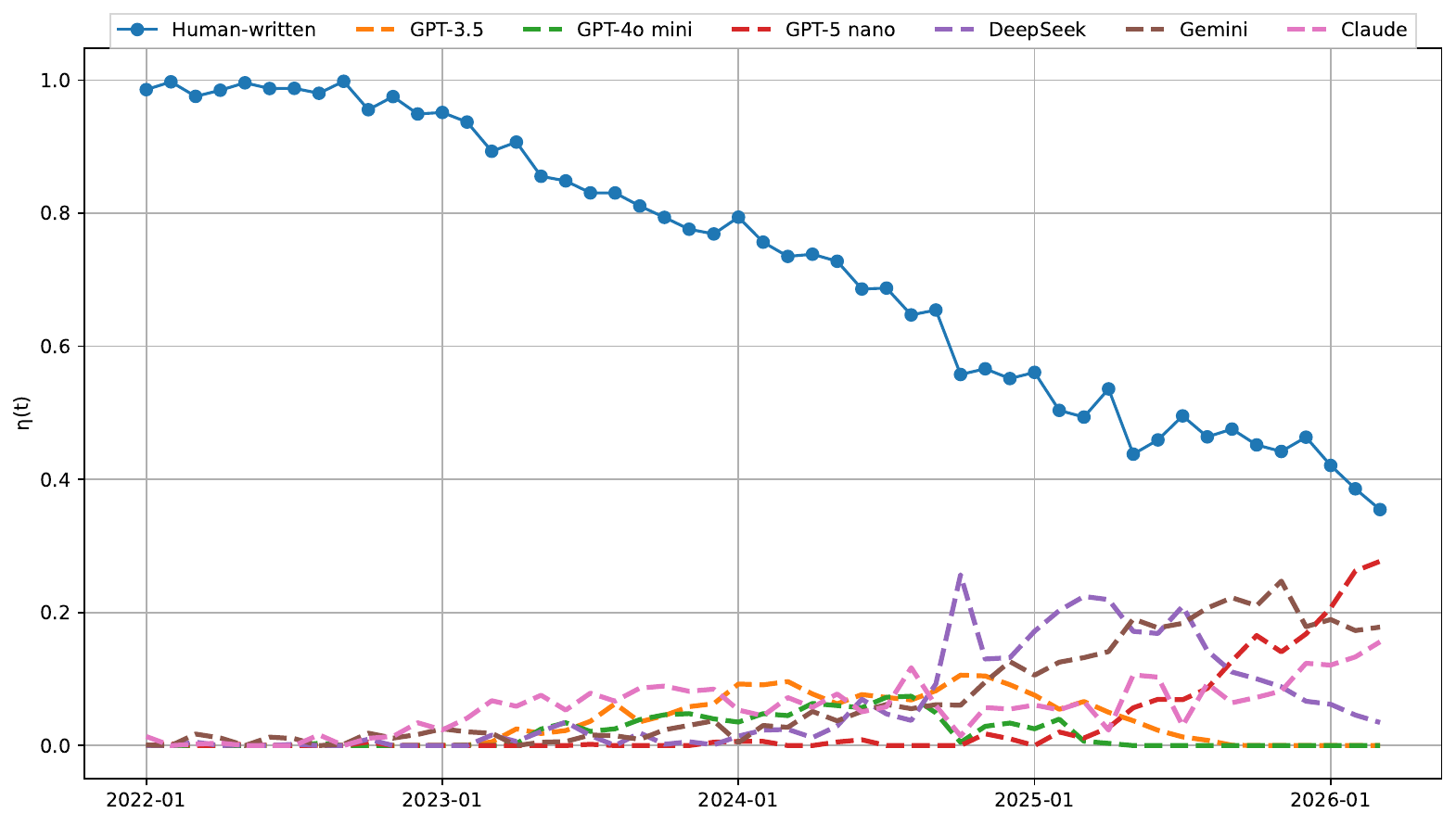}
        \caption{The 3,779 words used for estimation are drawn from the 10,000 most frequent words in the Google Books Ngram dataset that occur at least 5 times in the 2,000 preprocessed abstracts used for simulation.}
        \label{eta_estimation_5_0_10000}
    \end{subfigure}
    \hfill
    \begin{subfigure}{0.98\columnwidth}
        \centering
        \includegraphics[width=\textwidth]{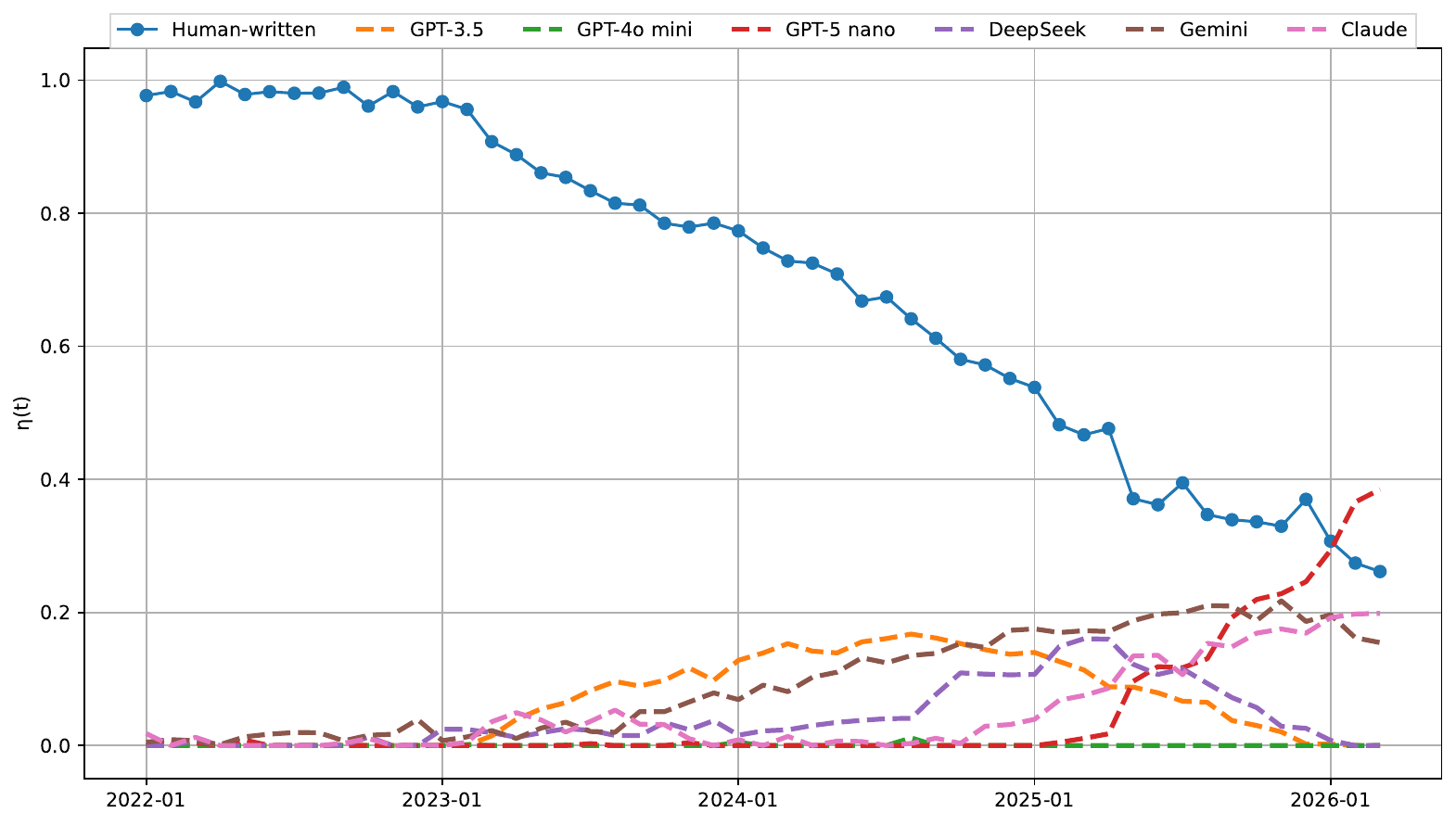}
        \caption{The 1,885 words used for estimation are drawn from the 10,000 most frequent words in the Google Books Ngram dataset that occur at least 20 times in the 2,000 preprocessed abstracts used for simulation.}
        \label{eta_estimation_20_0_10000}
    \end{subfigure}
    \caption{Supplementary results for the estimation of LLM impact in arXiv abstracts.}
    \label{eta_estimation_results_supplementary}
\end{figure*}

\end{document}